\documentclass[lettersize,journal]{IEEEtran}
\usepackage{amsmath,amsfonts}
\usepackage{algorithm}
\usepackage{array}
\usepackage{textcomp}
\usepackage{stfloats}
\usepackage{url}
\usepackage{verbatim}
\usepackage{graphicx}
\usepackage{cite}
\usepackage{multirow}
\usepackage{enumitem}
\usepackage{caption,subcaption}
\usepackage[switch]{lineno}
\usepackage{algpseudocode}
\usepackage{fancyhdr}
\usepackage{xcolor}
\usepackage{tikz}
\usepackage{comment}
\newcommand*{\circled}[2][]{\tikz[baseline=(C.base)]{
    \node[inner sep=0pt] (C) {\vphantom{1g}#2};
    \node[draw, circle, inner sep=1pt, yshift=1pt] 
        at (C.center) {\vphantom{1g}};}}
\begin{document}

\title{Attribute-Modulated Generative Meta Learning for Zero-shot Learning}

\author{Yun~Li,
        Zhe~Liu,
        Lina~Yao,
        and~Xiaojun~Chang
\thanks{Y. Li, Z. Liu, and L. Yao are with the School of Computer Science and Engineering, University of New South Wales, Sydney, NSW 2052, Australia (e-mail: yun.li5@student.unsw.edu.au;
zheliu912@gmail.com; lina.yao@unsw.edu.au). X. Chang is with the Faculty of Engineering and Information Technology, University of Technology Sydney, Sydney, NSW 2007, Australia (e-mail: xiaojun.chang@uts.edu.au)}
}


\maketitle

\begin{abstract}
Zero-shot learning (ZSL) aims to transfer knowledge from seen classes to semantically related unseen classes, which are absent during training.
The promising strategies for ZSL are to synthesize visual features of unseen classes conditioned on semantic side information and to incorporate meta-learning to eliminate the model's inherent bias towards seen classes.
While existing meta generative approaches pursue a common model shared across task distributions, we aim to construct a generative network adaptive to task characteristics.
To this end, we propose an \textbf{A}ttribute-\textbf{M}odulated gener\textbf{A}tive meta-model for \textbf{Z}ero-shot learning (AMAZ).
Our model consists of an attribute-aware modulation network, an attribute-augmented generative network, and an attribute-weighted classifier. 
Given unseen classes, the modulation network adaptively modulates the generator by applying task-specific transformations so that the generative network can adapt to highly diverse tasks. The weighted classifier utilizes the data quality to enhance the training procedure, further improving the model performance.
Our empirical evaluations on four widely-used benchmarks show that AMAZ outperforms state-of-the-art methods by 3.8\% and 3.1\% in ZSL and generalized ZSL settings, respectively, demonstrating the superiority of our method. Our experiments on a zero-shot image retrieval task show AMAZ's ability to synthesize instances that portray real visual characteristics.
\end{abstract}


\section{Introduction}

Object classification has undergone remarkable progress driven by the advances in deep learning. The underlying force ensuring the success is the availability of large amounts of carefully annotated image data. However, objects in the real world follow a long-tailed distribution~\cite{zhu2018generative,YangKCZCZ21}, i.e., a tremendous number of classes have few visual instances. Data insufficiency poses a bottleneck to the robustness of object classification methods. Targeting at overcoming this challenge, zero-shot learning has attracted plenty of interest recently~\cite{wang2018zero,verma2017simple,zhu2019generalized,gao2020zero,yang2020simple,verma2020meta,li2019zero}. 

Zero-shot learning (ZSL) aims to infer a classification model from \textit{seen classes}, i.e., classes with labeled samples that present in the training process, to recognize \textit{unseen classes}, i.e., classes that are absent from the training process.
It generally leverages semantic side information to transfer knowledge from seen classes to unseen classes. Typical side information include human-defined attributes that portray visual characteristics~\cite{farhadi2009describing,farhadi2010attribute,LuoCG21}, e.g., \textit{has tail}, and word embeddings of text descriptions~\cite{akata2015evaluation,pennington2014glove}.

A common strategy is to view ZSL as a visual-semantic embedding problem, which boils down to finding a projection that maps visual features and semantic features to the same latent space and performing nearest neighbor search in the space to predict labels~\cite{liu2018generalized,wang2019tafe,liu2020attribute,ye2019progressive,hu2020semantic,RenXCHLCW21}. However, this strategy suffers from the domain shift problem, due to distribution differences~\cite{fu2015transductive}. 
Some recent work uses generative methods to synthesize visual features conditioned on semantic side information and learn a conventional supervised classifier from generated samples to overcome the above issue~\cite{chen2018zero,zhu2019generalized,gao2020zero,felix2018multi,yang2020simple,XiaoLLCZC21}.
Given that generative models learned from seen classes exhibits inherent biases when generalizing to unseen classes,
meta generative approaches for ZSL emerge as a new trend to mitigate the biases~\cite{verma2020meta,meta_zeroshot1,meta_zeroshot2,meta_zeroshot3,meta_zeroshot4}. Meta generative approaches incorporate meta-learning models, e.g., Model-Agnostic Meta-Learning (MAML)~\cite{finn2017model}, into generative models. They divide seen classes into two disjoint sets (a support set and a query set) to mimic the ZSL setting and learn an optimal common generative model across seen and unseen classes.


\begin{figure}
    \centering
    \begin{subfigure}{0.23\textwidth}
    \centering
    \includegraphics[width=\textwidth]{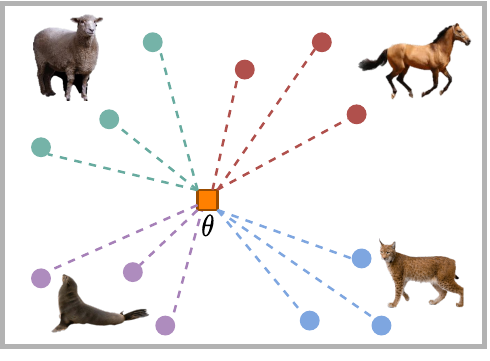}
    \caption{Globally-shared.}
    \end{subfigure}
    \begin{subfigure}{0.23\textwidth}
    \includegraphics[width=\textwidth]{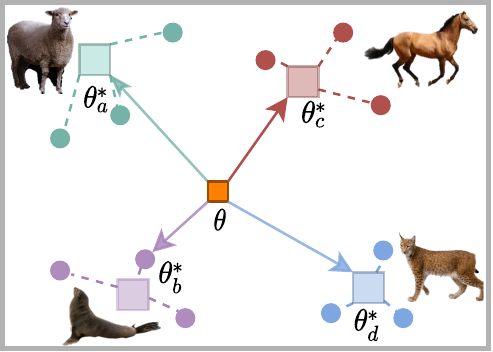}
    \centering
    \caption{Attribute-Modulated.}
    \end{subfigure}\hfil 
\caption{Visualization of model modulation. $\theta$ denotes the learned model representation. $\theta_a^*$, $\theta_b^*$, $\theta_c^*$ and $\theta_d^*$ represent modulated models for four tasks. (a) Globally-shared model is sub-optimal for a single task. (b) Attribute-modulated model adapts $\theta$  according to task characteristics, e.g., attribute embedding, to fit highly diverse tasks, which obtains task-specific models.}
\label{AMAZ_intro}
\end{figure}

\begin{figure*}[h]
    \centering
    \includegraphics[width=\linewidth]{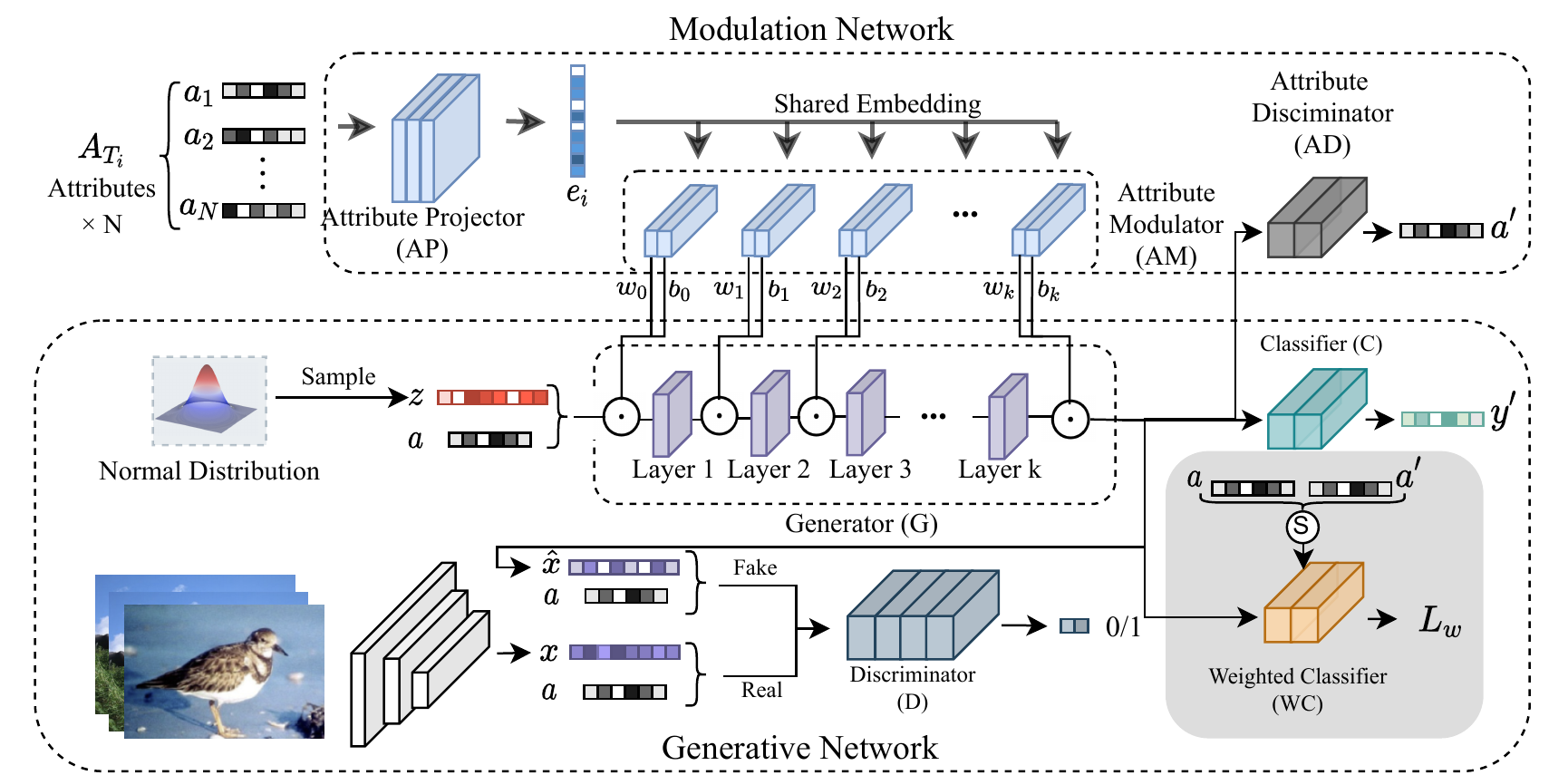}
    \caption{
    Model Architecture of AMAZ. AMAZ is composed of a modulation network, a generative network, and an attribute-weighted classifier to fulfill the final classification. The modulation network takes attribute sets as input to produce parameters that are further used to modify the generator. The $\bigodot$ denotes modulating operation. The \circled{S} calculates cosine similarity.}
    \label{fig AMAZ_model}
\end{figure*}

\textcolor{black}{Despite the effectiveness of meta generative approaches in ZSL, they still have limitations in learning characterized task distributions with diversities~\cite{sinha2021dibs}, e.g., in Computer Vision (CV)~\cite{david2020global} and Natural Language Processing (NLP)~\cite{reuver2021no}.} 
First, the common model shared across tasks may be sub-optimal when applied to a specific task~\cite{wang2020m, vuorio2019multimodal}; it may result in a deteriorated model, which seeks a common solution while neglecting individual tasks' characteristics.
For example, given four species (i.e., sheep, horse, seal, and bobcat) from the Animals with Attributes (AwA) dataset~\cite{lampert2013attribute}---which are highly diverse---if we consider each class as a task, then, as shown in Figure~\ref{AMAZ_intro} (a), the globally shared model representation $\theta$ learned by conventional meta generative methods would be far from being optimal for a single task.
In few-shot learning, using samples to fine-tune the model can help relieve the problem. But in ZSL, due to the lack of unseen data, the problem could be severe.
Second, the learned model is not guaranteed to generate samples that can simulate or reflect characteristics of unseen classes due to the absence of real images of unseen classes during training.
The synthetic data quality varies significantly across classes. The low-quality synthetic data may largely misguide and impair the training process of the final classifier.


In this paper, we generate features dynamically by proposing a novel \textbf{A}ttribute-\textbf{M}odulated gener\textbf{A}tive meta-model for \textbf{Z}ero-shot learning (AMAZ).
AMAZ can specialize a generalized model to adapt to diverse tasks, thus overcoming the first limitation.
Specifically, we augment the meta generative adversarial network with an attribute-aware modulation network, which modulates layers within the generator according to task characteristics, as illustrated in Figure~\ref{AMAZ_intro} (b). We propose an attribute discriminator for the modulation network to constrain its specialization direction. The specialization will be regularized towards the real data distribution. Moreover, we utilize the attribute discriminator to measure the synthetic data quality to modify the loss propagation of the weighted classifier. Thus, the weighted classifier can be robust to noisy low-quality data, which addresses the second limitation.

In summary, we make three-fold contributions:

\begin{itemize}
\item We propose an Attribute-Modulated generAtive meta-model for Zero-shot learning (AMAZ). AMAZ utilizes an attribute-aware modulation network to enhance the generative adversarial network and meta-learning. It combines the strengths of mitigating biases towards seen class and accommodating diverse tasks.

\item We introduce data quality to provide complementary guidance for a weighted classifier. The weighted classifier improves performance on all datasets, especially in SUN (3.6\%). 


\item We conduct extensive experiments in ZSL, Generalized-ZSL (GZSL), and zero-shot image retrieval tasks, where AMAZ consistently outperforms state-of-the-art algorithms on all four benchmarks, demonstrating our model's superiority. Our ablation studies and further analysis also testify to our model's robustness.

\end{itemize}

\section{Methodology}
\subsection{Problem Definition}
Let $D^{S}=\{(x, y, a)|x\in{X^{S}},y\in{Y^{S}},a\in{A^{S}}\}$ be the training data from seen classes, where $x\in{X^{S}}$ denotes the visual feature, $y\in{Y^{S}}$ denotes the class label of $x$, and $a\in{A^{S}}$ represents attributes (or any other kinds of semantic side information) of $y$. 
We define test data from unseen classes as $D^{U}=\{(x, y, a)|x\in{X^{U}},y\in{Y^{U}},a\in{A^{U}}\}$.
Seen and unseen classes are disjoint, i.e., $Y^{S}\cap{Y^{U}}= \emptyset$.
ZSL aims to learn a classifier $f_{ZSL}:x \in X^{S}$ that can classify objects from unseen classes. Generalized Zero-Shot Learning (GZSL) is more practical and challenging in that images from both seen and unseen classes may occur during the testing time: $f_{GZSL}:x\in X^{U}\cup X^{S}$.

In our model, we split $D^{\mathit{S}}$ into two disjoint sets $D^{S}_{\mathit{sup}}$ and $D^{\mathit{S}}_{\mathit{qry}}$ to function as support and query sets for meta-learning. We carry out episode-wise meta-training. In each episode, we sample task $\mathcal{T}_{i}=\{\mathcal{T}^{i}_{\mathit{sup}}, \mathcal{T}^{i}_{\mathit{qry}}\} \sim p(\mathcal{T})$ from $D^{\mathit{S}}_{\mathit{sup}}$ and $D^{S}_{\mathit{qry}}$, where $p(\mathcal{T})$ denotes the task distribution over $D^{S}$. $\mathcal{T}^{i}_{sup}$ and $\mathcal{T}^{i}_{qry}$ are sampled in $N$-way $K$-shot setting, which means that each $\mathcal{T}_{i}$ contains $N$ classes with $K$ labeled examples for each class. The number of $\mathcal{T}_{i}$ in an episode is decided by a hyper-parameter, i.e., batch size. We accumulate the gradients over all tasks in an episode for optimization.

The core of our proposed AMAZ (in Figure~\ref{fig AMAZ_model}) is an attribute-modulated meta-generative network that synthesizes task-specific visual features based on the attributes. Our goal is to learn a Generator (G) $f_{\theta_{g}}(a,z)\xrightarrow~\hat{x}$, where $\theta_{g}$ denotes the model parameters of G; $(a,z)$ denotes the given attribute and random noise; $\hat{x}$ denotes the synthesize features. We use meta-learning and modulation network to specialize the generalized model representation based on the current task information to better handle diverse tasks. AMAZ consists of three components: attribute-aware modulation network, attribute-augmented generative network, and attribute-weighted classifier.
The modulation network analyzes the attribute set of the current task to carry out task-specific model modulation.
The generative network is modulated by the modulation network to fit the current tasks better and thus synthesize more accurate visual features.
The weighted classifier measures the similarity of the synthesized features to enhance the training process of the final classifier.
The training of AMAZ is based on episode-wise meta-learning, which enables the learned model parameters to be more generalized than conventional ZSL methods~\cite{verma2020meta}. In the following sections, we will explain the model details and training procedures.

\subsection{Attribute-aware Modulation Network}
We use the attribute-aware modulation network to modulate the sub-optimal generator layer-wisely, then AMAZ can accommodate tasks with significant discrepancy.
For each task in the training episode, $\mathcal{T}_i$ consists of the objects from the current task. The attribute information can be denoted by $A_{\mathcal{T}_i}=\{a_{1},...,a_{N}\}$, which contains attributes corresponding to $N$ classes in $\mathcal{T}_{i}$. Then, we use an Attribute Projector (AP) to extract task embedding $e_i$ to represent the current task information:
\begin{equation}
    e_{i} = f_{\theta_{p}}(A_{\mathcal{T}_i})
\end{equation}
where $\theta_{p}$ denotes the model parameter of AP; $e_{i}$ denotes the representation of task information of $\mathcal{T}_i$; $A_{\mathcal{T}_i}$ denotes the corresponding attribute set of $\mathcal{T}_i$.

With the extracted task representation $e_{i}$, we use an Attribute Modulator (AM) to learn the modulation parameter $\{(w,b)\}$ that can adjust the layers in G:
\begin{equation}
(w_{j},b_{j})=h_{\theta_{m}}^{j}(e_{i})
\end{equation}
\begin{equation}\label{modulation operation}
\hat{o}_j=(1+Sigmoid(w_{j}))*o_{j}+Sigmoid(b_{j})
\end{equation}
where $o_{j}$ denotes the intermediate result of the $j^{th}$ layer in G; $h_{\theta_{m}}^{j}$ denotes the $j^{th}$ AM to modulate $o_{j}$; $(w_{j},b_{j})$ denotes modulation parameters for $o_{j}$; $\hat{o}_j$ denotes the modulated output.

Eq.~\ref{modulation operation} shows the detailed modulation operation. Since the task information is same for a certain task $\mathcal{T}_{i}$, we use the same input, i.e., $e_{i}$, for AMs. To produce different modulation parameters $\{(w,b)\}$, we use independent AMs for G, e.g., $j^{th}$ AM: $h_{\theta_{m}}^{j}\xrightarrow~\{(w_{j},b_{j})\}$ for $j^{th}$ layer.

\begin{figure}
\centering
\includegraphics[width=0.44\textwidth]{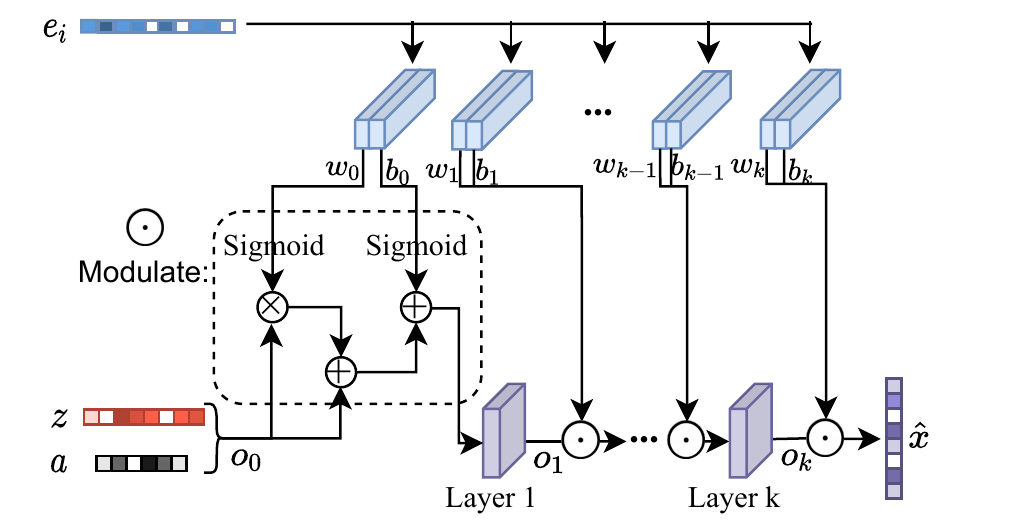}
\caption{Attribute-aware modulation. The detailed process of the modulating operation is circled by the dashed line. Modulation is performed layer by layer.}
\label{fig modulation}
\end{figure}

We take G with $k$ layers as an example and visualize the overall modulation flow in Figure~\ref{fig modulation}. We layer-wisely modulate all the intermediate outputs in G including the input $o_{0}=\{($z$, $a$)\}$ to obtain the final synthetic feature $\hat{x}$, which can gradually calibrate the model output to adapt to current tasks based on the extracted task representation.

To enable AP and AM to learn correct task embeddings and modulate G towards optimal directions, we design an Attribute Discriminator (AD):
\begin{equation}
    a'=f_{\theta_{ad}}(f_{\theta_{g}}(a,z))
\end{equation}
where $\theta_{ad}$ denotes the model parameters of AD; $a'$ is the reconstructed attribute.

We can view AD as a decoder, which attempts to reconstruct original input $a$. Therefore, we can compare $a'$ with $a$ to supervise the attribute-aware modulation network to improve the task information richness, which is consistent with the learning goals of AP and AM. Then, we can design the AD loss function $\mathcal{L}_{\mathcal{T}_i}^{AD}$ to optimize AP and AM for task $\mathcal{T}_{i}$:
\begin{equation}\label{loss_ad}
\begin{split}
\mathcal{L}_{\mathcal{T}_i}^{AD}&=\frac{1}{N*K}\sum_{n=1}^{N*K}\left \| a_{n}-a_{n}' \right \|^{2}_{2}\\
&= \frac{1}{N*K}\sum_{n=1}^{N*K}\left \| a_{n}-f_{\theta_{ad}}(f_{\theta_{g}}(a_{n},z)) \right \|^{2}_{2}
\end{split}
\end{equation}
where $\mathcal{T}_{i}$ contains $N*K$ samples; $a_{n}$ and $a_{n}'$ are the true attribute and the reconstructed attribute of $n^{th}$ sample in $\mathcal{T}_{i}$.

Similarly, by comparing the similarity of $a$ and $a'$, we can infer the quality of the generated features. In other words, we can use $a'$ to adjust the training of ZSL and GZSL classifiers to ease the influence of low-quality generation, which will be discussed in Section~\ref{weighted classifier}.



\subsection{Attribute-augmented Generative Network}
The attribute-augmented generative network consists of three components: an attribute-aware Generator (G) modulated by the modulation network to synthesize visual features: $\hat{x}\leftarrow f_{\theta_{g}}(z,a)$; an auxiliary Classifier (C) to categorize the input samples: $y'\leftarrow f_{\theta_{c}}(\hat{x})$; a Discriminator (D) to distinguish real or fake visual features: $\{0,1\} \leftarrow f_{\theta_{d}}(\hat{x},x)$.

For each task $\mathcal{T}_i$, following Eq.~\ref{modulation operation} and Figure~\ref{fig modulation}, given a random noise $z\in \mathbb{R}$ sampled from Gaussian distribution $\mathcal{N}(0,\sigma)$ and an attribute vector $a$, we can obtain the modulated synthetic visual features by $\hat{x}=f_{\theta_{g}}(z,a)$. Then, we design the loss function of D for $\mathcal{T}_i$ as $\mathcal{L}^{D}_{\mathcal{T}_{i}}$, which aims to optimize D to be able to distinguish fake features $\hat{x}\sim f_{\theta_{g}}(a,z)$ as 0 and real features $x\sim \mathcal{T}_{i}$ as 1:
\begin{equation}\label{loss_d}
\mathcal{L}^{D}_{\mathcal{T}_{i}}=\mathbb{E}_{a,x\sim\mathcal{T}_{i}}[f_{\theta_d}(x,a)]-\mathbb{E}_{a, z }[f_{\theta_d}(f_{\theta_{g}}(a,z),a))]
\end{equation}
where $\theta_{d}$ and $\theta_{g}$ denote the parameters of G and D, respectively; $z\sim \mathcal{N}(0,\sigma)$ denotes the random noise from normal distribution.




With the Eq.~\ref{loss_d}, we can obtain a reliable D to distinguish real and fake features. Then, we can optimize G to confuse D to synthesize more 'real' features. Besides from being real, we also need the classes of $\hat{x}$ to be easy to predict, and G can collaborate with modulation network to better suit $\mathcal{T}_{i}$. Therefore, we apply an auxiliary classifier C to enhance the class information richness of $\hat{x}$, and let $\mathcal{L}_{\mathcal{T}_i}^{AD}(a,a')$ be one of the learning goals of G. We design the loss function $\mathcal{L}^{GC}_{\mathcal{T}_{i}}$ for generator as follows:
\begin{equation}\label{loss_gc}
\begin{split}
\mathcal{L}^{GC}_{\mathcal{T}_{i}}=& \mathcal{L}^{G}_{\mathcal{T}_{i}}(a,z)+\mathcal{L}_{\mathcal{T}_i}^{AD}(a,a')+\mathcal{L}_{\mathcal{T}_i}^{CLS}(y',y) \\
=&-\mathbb{E}_{a, z\sim\mathcal{N}(0,\sigma))}[f_{\theta_d}(f_{\theta_{g}}(a,z),a)]+\mathcal{L}_{\mathcal{T}_i}^{AD}(a,a')\\
&+\mathcal{L}_{\mathcal{T}_i}^{CLS}(f_{\theta_{c}}(f_{\theta_{g}}(a,z)),y) 
\end{split}
\end{equation}
where $a'$ denotes the attribute vectors reconstructed by AD; $y$ is the true class label; $\mathcal{L}_{\mathcal{T}_i}^{CLS}$ is the classification loss measured using cross entropy. \textcolor{black}{Note that we use $\mathcal{L}_{\mathcal{T}_i}^{AD}$ in the optimization of AD to extract better attributes, but include it in $\mathcal{L}^{GC}$ to help GC generate semantic-rich samples.}

With Eq.~\ref{loss_d}-\ref{loss_gc}, we can construct a min-maxing loss function:
\begin{equation}\label{loss_zsl}
    \min_{\theta_{gc},\theta_{am}}\max_{\theta_{d}}\mathcal{L}^{D}_{\mathcal{T}_{i}} +\mathcal{L}^{GC}_{\mathcal{T}_{i}}
\end{equation}
where $theta_{gc}=\{\theta_{g},\theta_{c}\}$ is the parameters of G and C; $\theta_{am}=\{\theta_{p},\theta_{M},\theta_{ad}\}$ is the parameters of AP, AMs, and AD; $\theta_{M}=\{\theta_{m}^{j}:j\in[1,k]\}$ is the set of AMs; $\theta_{d}$ is the parameters of D.

We optimize Eq.~\eqref{loss_zsl} in an adversarial manner through the following: 1) maximizing $\mathcal{L}^{D}_{\mathcal{T}_{i}}$ to enable $D$ to distinguish between real or generated samples;
2) optimizing $\theta_{am}$ to minimize $\mathcal{L}_{\mathcal{T}_i}^{GC}$ to enhance the quality of generated samples;
3) optimizing $\theta_{gc}$, i.e., the generator and classifier, to minimize $\mathcal{L}^{GC}_{\mathcal{T}_{i}}$ to fool the discriminator and assist attribute modulation network. 

\subsection{Meta-training Procedure}

\begin{algorithm}[h]
  \caption{AMAZ Training Procedure}\label{algorithm_process}
  \begin{algorithmic}[1]
   \Require $p(\mathcal{T})$: task distribution,$D^S$:training dataset
   \Require $\alpha_{1},\alpha_{2},\alpha_{3},\beta_{1},\beta_{2},\beta_{3}$, batch size 
   \State Initialize $\theta_{d},\theta_{gc},\theta_{am}$
   \While {not DONE do}
   \State Split $\mathcal{D}^{S}$ into disjoint subsets $\mathcal{D}^{S}_{sup}$ and $\mathcal{D}^{S}_{qry}$
   \State Sample batches of tasks $\mathcal{T}_{i}=\{\mathcal{T}^{i}_{sup}, \mathcal{T}^{i}_{qry}\} \sim p(\mathcal{T})$ over $D^{S}_{sup}$ and $D^{S}_{qry}$ respectively
   \For{all $i$}
   \State Compute $e_{i} \leftarrow f_{\theta_{p}}(A_{\mathcal{T}_i})$ with $N$ classes in $\mathcal{T}^{i}_{sup}$
   \For{all $j$}
   \State Generate $(w_{j},b_{j})\leftarrow h_{\theta_{m}^{j}}(e_{i})$ to modulate G
   \EndFor
   \State Evaluate $\bigtriangledown_{\theta_{d}}\mathcal{L}_{\mathcal{T}^{i}_{sup}}^{D}(\theta_{d})$ w.r.t. samples in $\mathcal{T}^{i}_{sup}$
   \State Evaluate $\bigtriangledown_{\theta_{am}}\mathcal{L}_{\mathcal{T}^{i}_{sup}}^{AD}(\theta_{am})$ w.r.t. samples in $\mathcal{T}^{i}_{sup}$
   \State Update $\theta_{d}^{‘} \leftarrow \theta_{d}+\alpha_{1}\bigtriangledown_{\theta_{d}}\mathcal{L}_{\mathcal{T}^{i}_{sup}}^{D}(\theta_{d})$
   \State Update $\theta_{am}^{’} \leftarrow \theta_{am}-\alpha_{2}\bigtriangledown_{\theta_{am}}\mathcal{L}_{\mathcal{T}^{i}_{sup}}^{AD}(\theta_{am})$
   \State Evaluate $\bigtriangledown_{\theta_{gc}}\mathcal{L}_{\mathcal{T}^{i}_{sup}}^{GC}(\theta_{gc})$ w.r.t. samples in $\mathcal{T}^{i}_{sup}$
   \State Update $\theta_{gc}^{'} \leftarrow \theta_{gc}-\alpha_{3}\bigtriangledown_{\theta_{gc}}\mathcal{L}_{\mathcal{T}^{i}_{sup}}^{GC}(\theta_{gc})$
  \EndFor
  \State Update $\theta_{d}\leftarrow\theta_{d}+\beta_{1}\sum_{\mathcal{T}^{i}_{qry}}\bigtriangledown_{\theta_{d}}\mathcal{L}_{\mathcal{T}^{i}_{qry}}^{D}(\theta_{d}^{'})$
  \State Update $\theta_{am}\leftarrow\theta_{am}-\beta_{2}\sum_{\mathcal{T}^{i}_{qry}}\bigtriangledown_{\theta_{am}}\mathcal{L}_{\mathcal{T}^{i}_{qry}}^{AD}(\theta_{am}^{'})$
  \State Update $\theta_{gc}\leftarrow\theta_{gc}-\beta_{3}\sum_{\mathcal{T}^{i}_{qry}}\bigtriangledown_{\theta_{gc}}\mathcal{L}_{\mathcal{T}^{i}_{qry}}^{GC}(\theta_{gc}^{'})$
\EndWhile
  \end{algorithmic}
\end{algorithm}

Following~\cite{finn2017model}, we conduct episode-wise meta-training for our model in a model-agnostic manner (described in Algorithm~\ref{algorithm_process}).
In each iteration, we first sample tasks $\mathcal{T}_{i}=\{\mathcal{T}^{i}_{sup},\mathcal{T}^{i}_{qry}\} \sim p(\mathcal{T})$, where $\mathcal{T}^{i}_{sup}$ and $\mathcal{T}^{i}_{qry}$ are sampled over $D^{S}_{sup}$ and $D^{S}_{qry}$, respectively (line 4). Considering a pre-split of training dataset
could restrict the effectiveness of task sampling, in our experiment, we sample tasks from the whole training set---we only ensure the training and validation classes in each task are disjoint (line 3). \textcolor{black}{After selecting classes for each task, we randomly sample images from these classes to construct support and query sets.}
Since $\mathcal{T}^{i}_{sup}$ and $\mathcal{T}^{i}_{qry}$ are disjoint to mimic seen and unseen classes in the ZSL setting, our AMAZ can learn to generate samples for unseen classes by transferring knowledge from seen classes.

\textcolor{black}{Next, for each task, we modulate the generator according to the attribute set (lines 6 - 8). Then, $\mathcal{T}^{i}_{sup}$ is used for fast task adaptation (lines 10 - 15). We first optimize $\mathcal{L}^{D}$ and $\mathcal{L}^{AD}$ on support set to find the task-specific parameters for each task (lines 12 - 13), and then with updated $D$ and $AD$, we can optimize $L^{GC}$ on support set (lines 15):}
\textcolor{black}{
\begin{equation}
    \theta^{'} \leftarrow \theta-\alpha\bigtriangledown_{\theta}\mathcal{L}_{\mathcal{T}^{i}_{sup}}^{D}(\theta)
\end{equation}
}
\textcolor{black}{Note that D, AM, and GC are updated with different learning rate. Specifically, for D, the equation is $\theta^{'} \leftarrow \theta+\alpha\bigtriangledown_{\theta}\mathcal{L}_{\mathcal{T}^{i}_{sup}}^{D}(\theta)$. We use $-$ for AM and GC to minimize $L^{GC}$ and $\mathcal{L}^{AD}$, and use $+$ for D to maximize $\mathcal{L}^{D}$. }

\textcolor{black}{Given the gradient from tasks in support set, we further use $\mathcal{T}^{i}_{qry}$ to update meta parameters, which are shared across all tasks, to obtain better generalized parameters (lines 17 - 19):}

\textcolor{black}{
\begin{equation}
    \theta\leftarrow\theta-\beta\sum_{\mathcal{T}^{i}_{qry}}\bigtriangledown_{\theta}\mathcal{L}_{\mathcal{T}^{i}_{qry}}(\theta^{'})
\end{equation}
}

\textcolor{black}{Also, all modules are updated with different learning rate, and $\mathcal{L}^{D}$ is optimized in direction opposite $\mathcal{L}^{AD}$ and $\mathcal{L}^{GC}$.}

Different from MAML, during fast task adaptation, parameters are updated per batch-of-tasks instead of per task to enhance robustness~\cite{verma2020meta}.

\subsection{Attribute-weighted Classifier}\label{weighted classifier}
In this section, we first introduce weighted classifier based on the reconstructed attribute $a'$ from AD, and then introduce the inference process of ZSL/GZSL.

\textbf{Weighted classifiers}. ZSL/GZSL aims to train a classifier based on synthetic features to predict seen or unseen classes. Considering that the data quality of synthetic features differs from each other, we can adjust the loss weight based on the data quality to train the classifier to better fit the true data distribution and prevent fitting unreal features. We use cosine similarity between $a$ and $a'$  to measure the data quality:
\begin{equation}
    Q(a')=cos(a,a')=\frac{a\cdot a'}{\left \| a \right \|\left \| a' \right \|}
\end{equation}
where $Q(a')$ denotes the data quality of $a'$; $\cdot$ denotes dot product; $\left \| a \right \|$ and $\left \| a' \right \|$ denote magnitude of $a$ and $a'$.

Then, we propose two weighted classifier based on Softmax (fully-connected layers followed by a Softmax layer) and Linear-SVM, respectively. For Softmax classifier which is trained in batch-wise, we can directly measure the instance-level data quality to adjust the gradient propagation to obtain more accurate classification loss:
\begin{equation}
    \mathcal{L}_{soft}=\frac{1}{|A'|}\sum_{a'\in A'}Q(a',a)\cdot CE(Softmax(a'))\phi(a))
\end{equation}
where $a'\in A'$ denotes the generated features from current task; $a$ denotes the corresponding true attribute of $a'$; $|A'|$ denotes the sample number of $A'$; CE denotes cross entropy loss; $Softmax(a')\rightarrow~\hat{y}$ denotes the predicted label; $\phi(a)\rightarrow~y$ denotes the corresponding true class label.

For Linear-SVM which is trained in dataset-wise, we propose a more efficient weighted loss function via class-level weights. We calculate class weights:
\begin{equation}\label{SVM class weights}
    w_{y}=\frac{1}{|A_{y}|}\sum_{a'\in A_{y}}Q(a',a)
\end{equation}
where $A_{y}=\{a_{i}':\phi(a_{i}')=y\}$ denotes the set of $a'$ whose true class label is $y$; $|A_{y}|$ denotes the sample number in $A_{y}$. Thus, we can calculate the weights for classes as the class weights for Linear-SVM to adjust the learning process.

\textbf{Inference process}. Given an unseen class, AMAZ first replicates its attribute $N$ times to construct attribute set, and then embeds the attribute set to produce attribute-aware parameters for modulation of the well-trained generator.
Then, it applies the customized task-specific generator to synthesize visual features. 
By sampling $z$, an arbitrary number of visual features can be generated. With the generated visual features, a conventional supervised classifier, e.g., Linear-SVM and Softmax, can be trained to solve ZSL, i.e., the classifier only trained with synthesized features can be used to classify real features. 
In GZSL, samples for both seen and unseen classes are generated to train the final classifier. We use synthesized features instead of real data for seen classes to avoid bias.

\section{Experiments}

\subsection{Experiment Setup}
\textbf{Datasets:} We conduct a comprehensive evaluation of our method in ZSL and GZSL settings on four widely used benchmark datasets: SUN~\cite{patterson2012sun}, CUB~\cite{welinder2010caltech}, AWA1~\cite{lampert2009learning}, and AWA2~\cite{xian2019zero}. 
CUB and SUN are datasets of bird species and scenes, respectively; both are considered challenging since they are fine-grained and each class has limited data. 
AWA1 and AWA2 are coarse-grained datasets, where images come from highly diverse animals.

We adopt the commonly used 2048-dimensional CNN features extracted by ResNet101~\cite{he2016deep} as visual features and use pre-defined attributes as semantic side information except for the CUB dataset. For CUB, we follow~\cite{verma2020meta} to use CNN-RNN textual features as semantic information~\cite{reed2016learning}, which perform superior to the hand-hand-engineered attributes.
Moreover, the datasets are divided into seen and unseen classes following the commonly used Proposed Split (PS)~\cite{xian2019zero}, and all the competitors use this split.
Table~\ref{table datasets} shows the statistics and splits of the datasets.

\begin{table}
\centering
\caption{Statistics of experimental datasets}
\label{table datasets}
\begin{tabular}{l|ccccc}
\hline
Datasets & Attribute dim & Image num& Seen/Unseen classes  \\ 
\hline
\hline
AWA1 & 85 & 30475 & 40/10\\
AWA2 & 85 & 37322 & 40/10\\
CUB & 1024 & 11788 & 150/50\\
SUN & 102 & 14340 & 645/72\\
\hline
\end{tabular}
\end{table}

\textbf{Implementation Details:} 
We sample the support and query sets in 10-way-5-shot and 10-way-3-shot, respectively, and \textcolor{black}{each batch has 10 tasks, i.e., each batch processes 800 images. The model is first optimized based on the 500 support images, and then the gradients are calculated by accumulating validation loss of the 300 images in query sets}.  We set the learning rates in Algorithm~\ref{algorithm_process}: $\alpha_{1}=\alpha_{2}=\alpha_{3}=1e-3$, $\beta_{1}=1e-3,\beta_{2}=\beta_{3}=1e-5$. We use Adam optimizer to train the model. The epoch number is 25000 for SUN and CUB, and 15000 for AWA1 and AWA2. \textcolor{black}{We set a larger number for SUN/CUB than for AWA1/2 because SUN/CUB has over 200 classes, while AWA1/2 only has 50 classes. We set the numbers to allow full training and report the final results.} We set $\sigma$ to different values for training ($\sigma=0.1$) and testing ($\sigma=1$) to prevent the generator from being biased towards seen classes during testing.
During testing, we consider the single class as a task and duplicate its attribute to modulate the generator first. Then, we use the modulated generator to synthesize visual features. We generate 100 samples for each unseen class in ZSL and 300 samples for both seen and unseen classes in GZSL. Network modules are implemented by the multi-layer perception. We adopt Dropout layer with parameter 0.5, Batch Normalize (BN) layer with parameter 0.8 and LeakyReLU (LeReLU) with parameter 0.5 as activation function in our network. 

\begin{table}[htb]
\centering
\caption{Overall comparison in ZSL. The performance is evaluated by average per-class Top-1 accuracy (\%). Non-generative and generative methods are listed at the top and bottom, respectively. We embolden the best result and underline the second-best result for each dataset.}
\label{table zsl}
\begin{tabular}{l|c|c|c|c}
\hline
Method & ~SUN~& ~CUB~ & ~AWA1~ & ~AWA2~  \\ 
\hline
\hline
ESZSL  \cite{romera2015embarrassingly} & 54.5 & 53.9& 58.2 & 58.6\\
LATEM \cite{xian2016latent} & 55.3 & 49.3  & 55.1 & 55.8 \\
SAE  \cite{kodirov2017semantic} & 59.7 & 50.9 &53.0 &66.0\\
RelationNet \cite{sung2018learning}  & - &55.6&68.2&64.2\\
PREN  \cite{ye2019progressive} & \underline{60.1} & 61.4 & - & 66.6 \\
SGV-18  \cite{hu2020semantic} & 59.0 & 67.2 & - & 67.5 \\

\hline 
\hline
VZSL  \cite{wang2018zero} & 59.0 & 56.3 & 67.1 & 66.8\\
MCGZSL  \cite{felix2018multi} & 60.0 & 58.4 & 66.8 &67.3\\
FGZSL  \cite{xian2018feature} & 58.6 & 57.7& 65.6 & 68.2   \\
TVN  \cite{zhang2019triple} & 59.3 & 54.9 & 64.7 & -\\
Zero-VAE-GAN \cite{gao2020zero} & 58.5 & 51.1 & 68.5 & 66.2\\
SELAR-GMP  \cite{yang2020simple} & 58.3 & 65.0 & - & 57.0   \\
MM-WAE \cite{chen2020generalized} & 58.2 & 55.0 & 65.2 & 65.5 \\
ZSML (baseline)\cite{verma2020meta}  & 57.9 & 68.3 & 67.3 & 68.6  \\
\hline
\hline
AMAZ softmax (ours) & 57.1 & 66.6& 67.5 & 67.6   \\
AMAZ weighted-soft (ours) & \textbf{60.7} & 68.9& 68.1 & 68.2  \\
AMAZ svm (ours) & 59.4 & \underline{69.6} & \underline{71.7} & \textbf{72.4 }  \\
AMAZ weighted-svm (ours) & 59.7 & \textbf{70.0} & \textbf{71.9} & \underline{71.7 } \\
\hline
\end{tabular}
\end{table}

\begin{table*}[htb]
\centering
\caption{Overall comparison in GZSL. The performance is evaluated by average per-class Top-1 accuracy (\%) on seen classes(S), unseen classes (U), and their harmonic mean (H). We embolden the best result and underline the second-best result on each dataset.}
\label{table gzsl}
\begin{tabular}{l|ccc|ccc|ccc|ccc}
\hline
\multirow{2}{*}{Method} & \multicolumn{3}{c}{AWA2} & \multicolumn{3}{c}{AWA1} & \multicolumn{3}{c}{SUN} & \multicolumn{3}{c}{CUB} \\ 
\cline{2-13}
& U & S & H & U & S & H & U & S & H & U & S & H \\
\hline
\hline
ESZSL  \cite{romera2015embarrassingly} & 5.9 & 77.8 & 11.0 & 6.6 & 75.6 & 12.1 & 11.0 & 27.9& 15.8& 12.6& 63.8 &21.0 \\
SYNC  \cite{changpinyo2016synthesized}& 10.0 & 90.5 & 18.0 &8.9 & \textbf{87.3} & 16.2 & 7.9& \textbf{43.3}& 13.4& 11.5 & \underline{70.9} & 19.8 \\
DEM  \cite{zhang2017learning} & 30.5 & 86.4 & 45.1 & 32.8 & \underline{84.7} & 47.3 & 20.5& 34.3& 25.6& 19.6 & 57.9 & 29.2 \\
Gaussian-Kernal  \cite{zhang2018zero} & 18.9 & 82.7 & 30.8 & 17.9 & 82.2 & 29.4 & 20.1 & 31.4 & 24.5 & 21.6 & 52.8 & 30.6\\
TAFE-Net  \cite{wang2019tafe} &  36.7 & \underline{90.6} & 52.2 & 50.5 & 84.4 & 63.2 & 27.9 & \underline{40.2} & 33.0 & 41.0 & 61.4 & 49.2 \\
PQZSL  \cite{li2019compressing} & 31.7 &70.9 &43.8&-&-&-&35.1 &35.3 &\underline{35.2}& 43.2 &51.4 &46.9\\
\hline
\hline
SP-AEN  \cite{chen2018zero} & 23.0 & \textbf{90.9} & 37.1 & - & - & -& 24.9&38.6 & 30.3& 34.7 & 70.6 & 46.6  \\
GAZSL  \cite{zhu2018generative} & 35.4 & 86.9 & 50.3 &29.6 & 84.2 & 43.8 &22.1 & 39.3 & 28.3 &31.7 & 61.3 & 41.8 \\
TVN  \cite{zhang2019triple} & - & - & -& 27.0 & 67.9 & 38.6 & 22.2 & 38.3 & 28.1 & 26.5 & 62.3 & 37.2 \\
Zero-VAE-GAN  \cite{gao2020zero} & 51.7 & 74.8 & 61.1 & 50.5 & 67.8 & 57.9 & \textbf{49.0} & 26.0 & 34.0 & 40.5 & 47.8 & 43.9\\
SELAR-GMP  \cite{yang2020simple}  & 32.9 & 78.7 & 46.4 & - & - &- & 23.8 & 37.2 & 29.0 & 43.0 & \textbf{76.3} & \underline{55.0} \\
ZSML  \cite{verma2020meta}  & 51.6 & 75.6 & 61.4 & 52.5 & 69.2 & 59.7 & 25.2& 35.9& 29.6& \underline{48.6} & 60.1 & 53.7 \\
\hline
\hline
AMAZ weighted-soft (ours)  & \textbf{60.1} & 69.2 &\textbf{64.3} & \textbf{64.4} & 63.6 & \underline{64.1} & \underline{42.0}& 35.1&\textbf{38.3} & \textbf{58.2}& 55.7& \textbf{56.9}\\
AMAZ weighted-svm (ours)  & \underline{56.0}& 74.6&\underline{64.0}& \underline{57.6}&75.5 & \textbf{65.3}& -& -& -& -& -& -\\
\hline
\end{tabular}
\end{table*}

\subsection{Zero-shot Learning}
Table~\ref{table zsl} shows the results of our experimental comparison with two groups (non-generative and generative) of methods, i.e., 14 state-of-the-art algorithms, in ZSL.
We use the results reported in original papers or summarized in previous work~\cite{zhu2019learning} in the table.
 Note that the results of ZSML in~\cite{verma2020meta} are evaluated by overall accuracy; thus we re-run ZSML~\cite{verma2020meta} in our environment and utilize average per-class Top-1 accuracy as the evaluation metric to ensure a fair comparison. 
We report our results using Linear-SVM, Softmax (i.e., two fully-connected layers followed by one Softmax layer), and their attribute-weighted versions, i.e., weighted-soft and weighted-svm, as final classifiers, respectively.

AMAZ consistently outperforms state-of-the-art methods and achieves 0.6\%, 1.7\%, 3.4\%, and 3.8\% improvements than the second-best method on SUN, CUB, AWA1, and AWA2, respectively.
Besides, AMAZ exhibits more significant improvement on AWA1 and AWA2 than on SUN and CUB. It is reasonable because the objects are coarse-grained in AWA1 and AWA2 and have greater differences than the objects in SUN and CUB.

\textbf{Component ablation study.} 
Considering that ZSML is also a generative model trained by MAML, we take ZSML as our baseline.
We compare ZSML and AMAZ adopting 4 different classifiers as ablations in ZSL. Our proposed attribute modulation network is effective on four benchmark datasets, demonstrated by the improvements by up to 2.8\%, 1.7\%, 4.6\%, and 3.8\% on SUN, CUB, AWA1, and AWA2, respectively, when compared with ZSML. Also, the superiority of the weighted-versions of SVM and Softmax proves that our proposed attribute-weighted loss, both instance-level and class-level, can effectively improve the performance of the final classifier. Besides, there exists a performance gap between the SVM-based and Softmax-based classifiers. The reason lies in that we only train the Softmax-based classifier for 20 epochs to avoid gaining improvements from a better-trained classifier.

\subsection{Generalized Zero-shot Learning}

We follow~\cite{xian2019zero} and calculate the average per-class Top-1 accuracy on seen (denoted by $S$) and unseen classes (denoted by $U$), and their harmonic mean (defined as $H=\frac{2*U*S}{U+S}$) to evaluate the performance in the GZSL setting.
Table~\ref{table gzsl} shows the results comparing AMAZ with state-of-the-art methods. SVM is time and space consuming when classifying over 200 classes; thus we only report weighed-svm results for AWA2 and AWA1.

As shown in Table~\ref{table gzsl}, AMAZ surpasses all the other approaches and achieves 2.9\%, 2.1\%, 3.1\%, and 1.9\% improvements in harmonic mean on AWA2, AWA1, SUN, and CUB, respectively. Also, our model performs best on the unseen classes of AWA1, AWA2, and CUB, and second-best on SUN.
This implies that our model can effectively infer visual features for unseen classes and eliminate bias towards seen classes. The improvements derive from three aspects: 1) the incorporation of meta-learning; 2) the use of attribute modulation network to make model better adapting to diverse tasks; 3) the guidance of attribute-weighted loss in training final classifiers to denoise low-quality generated data. 

\begin{figure}[htb]
\centering 
\begin{subfigure}{0.23\textwidth}
\centering
\includegraphics[width=\textwidth]{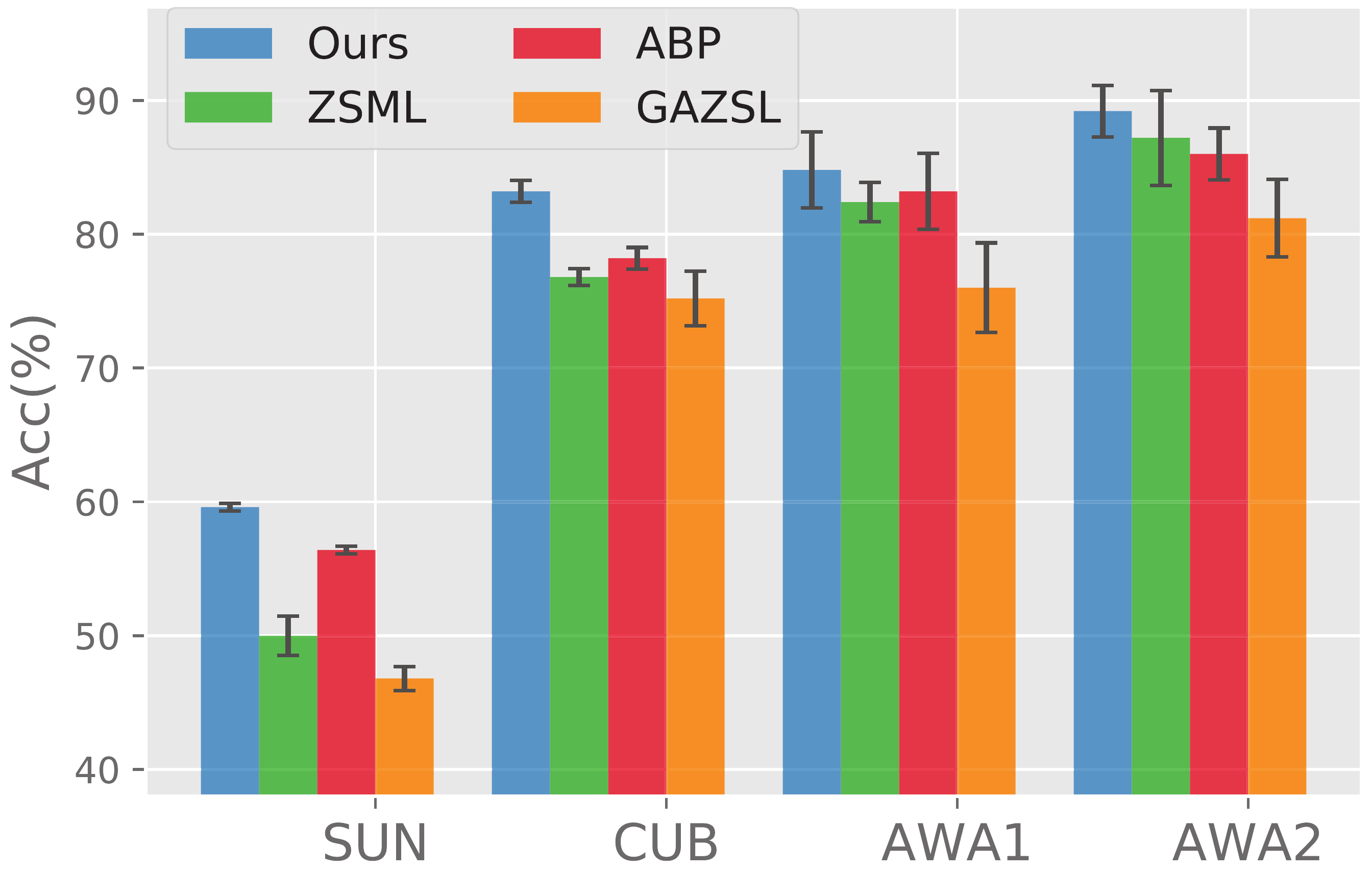}
\caption{Top-5 Image retrieval.}
\end{subfigure}\hfil 
\begin{subfigure}{0.23\textwidth}
\centering
\includegraphics[width=\textwidth]{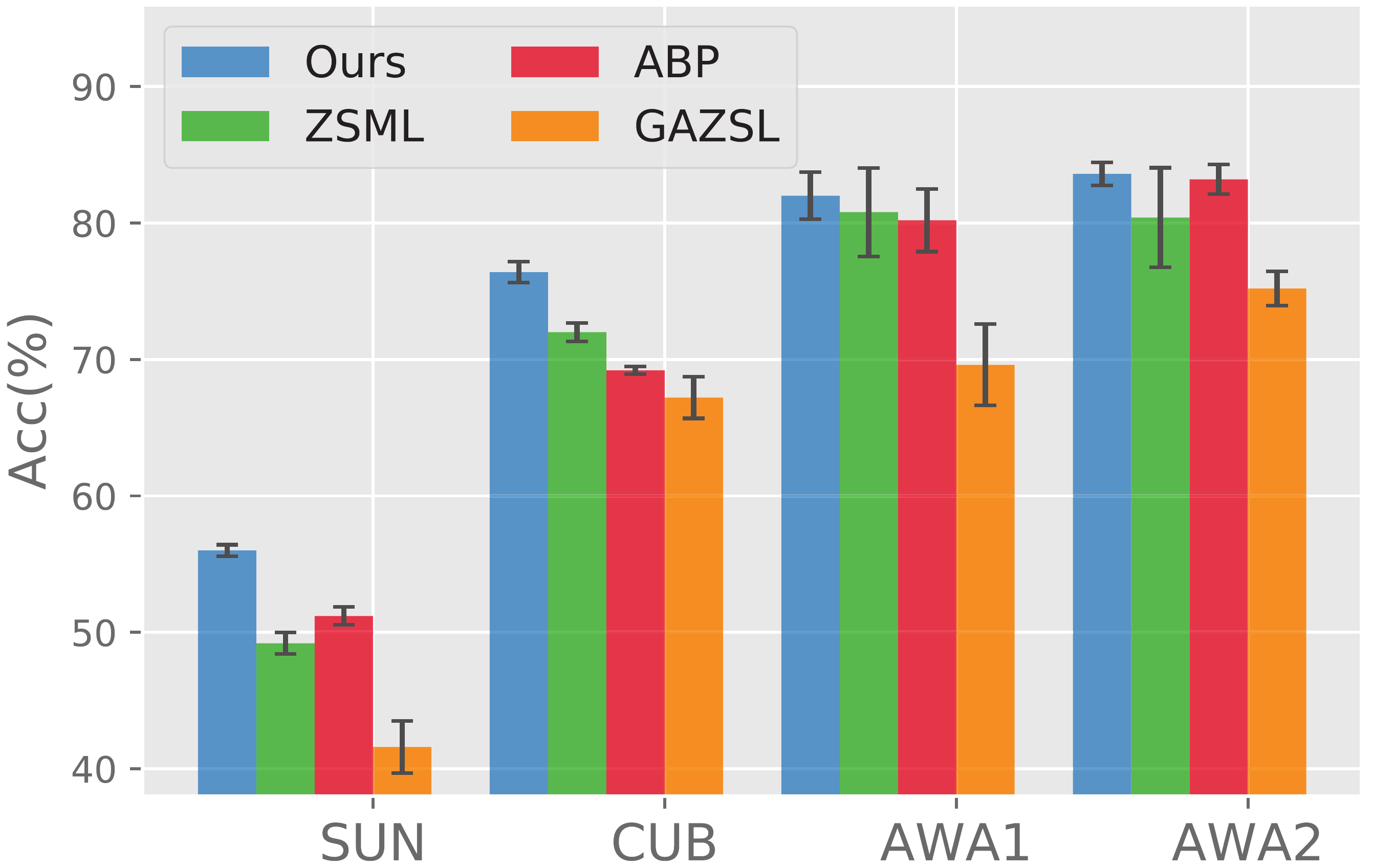}
\caption{Top-10 Image retrieval. }
\end{subfigure}\hfil 

\caption{Zero-shot image retrieval in average precision (\%).}
\label{fig retrieval}
\end{figure}

\begin{figure*}[htb]
    \centering
    \includegraphics[width=\linewidth]{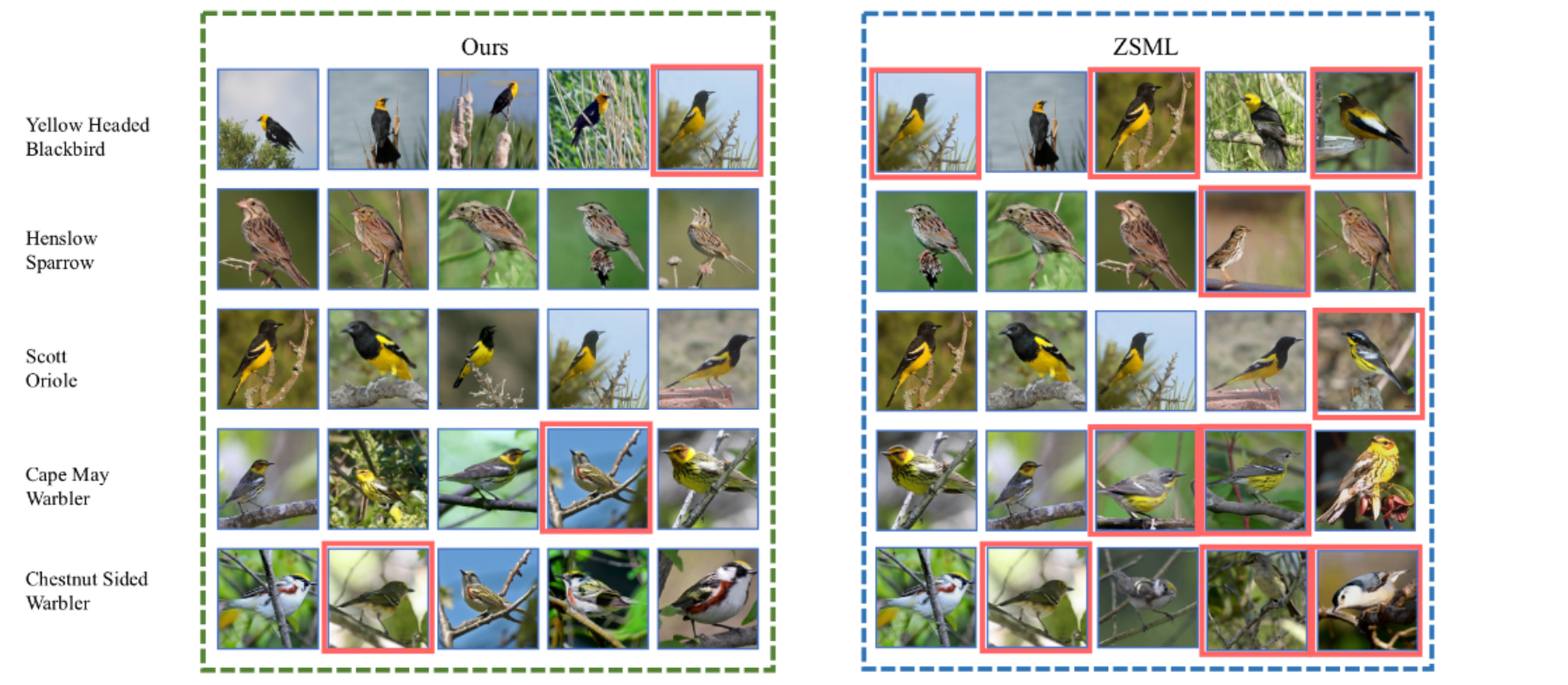}
    \caption{
    Visualization of image retrieval on CUB. Each row contains the five retrieved images given attributes of a specific class. The images circled by pink boxes are wrong results.
    }
    \label{fig image_retrieval}
\end{figure*}

\subsection{Zero-shot Image retrieval}

Given the attributes of unseen classes, the zero-shot image retrieval problem aims to retrieve the most relative images. 
We compare our model with three generative models, i.e., GAZSL~\cite{zhu2018generative}, ZSML~\cite{verma2020meta}, and ABP~\cite{zhu2019learning}, for the zero-shot image retrieval task on four datasets.
We adopt two settings: retrieving the Top-5 and Top-10 images for each class from the whole dataset.
Specifically, given the attributes, we utilize the generator to synthesize 100 features, calculate the average values of the synthetic features as representatives, and then retrieve the images that are nearest to the representatives from real unseen data. 

Figure~\ref{fig retrieval} shows the results evaluated by average precision.
AMAZ outperforms competitors by up to 5\% in two settings, demonstrating that the attribute modulation network can enhance the generative quality. Besides, our model is more stable than competitors, reflected by minor error bars.

We also provide qualitative results of using ZSML and AMAZ to retrieve five images that are most likely belonging to five given bird species from CUB.
Fig~\ref{fig image_retrieval} shows that both ZSML and AMAZ  perform well in `Henslow Sparrow', which is easy to distinguish.
AMAZ shows great superiority over ZSML in highly similar species, i.e., `Yellow-headed Blackbird' vs. `Scott Oriole', and `Cape May Warbler' vs. `Chestnut Sided Warbler'.
The possible reason is that ZSML, as a GAN-based model, is prone to mode collapse when CUB contains 200 fine-grained classes. In contrast, AMAZ, by introducing attribute discriminator and attribute loss, can generate highly discriminative features and achieve more inter-class diversity, thus avoiding mode collapse.

Overall, our model consistently outperforms competitors in different settings (as shown in Tables \ref{table zsl} and \ref{table gzsl}), and different tasks (as shown in Figures~\ref{fig retrieval} and~\ref{fig image_retrieval}), demonstrating the robustness and the superiority of AMAZ.

\begin{table}[htb]
\caption{Comparisons of the attribute-aware modulator operation. The performance is evaluated by average per-class Top-1 accuracy ($\%$). \textit{w/o} indicates without, and \textit{w} the opposite.}
\label{table modular}
\begin{tabular}{c|c|c|c|c}
\hline
 ~~~\textit{base}~~~& ~~~\textit{operator}~~~ & ~~~\textit{activation}~~~ & ~~~\textit{bias}~~~&~~~Acc~~~  \\ 
\hline
\hline
w/o&w/o&w/o&w/o&10.8\\
w/o&w/o&\textit{Sigmoid}&w/o&22.9\\
w/o&w/o&\textit{Softmax}&w/o&22.5\\
w&$+$&\textit{Sigmoid}&w/o&\underline{72.0}\\
w&$-$&\textit{Sigmoid}&w/o&71.2\\
w&$+$&\textit{Softmax}&w/o&70.8\\
w&$-$&\textit{Softmax}&w/o&69.6\\
w&$+$&\textit{Sigmoid}&w&\textbf{72.4}\\
\hline
\end{tabular}
\end{table}

\begin{figure*}[htb]
\centering 
\begin{subfigure}{0.23\textwidth}
\centering
\includegraphics[width=\textwidth]{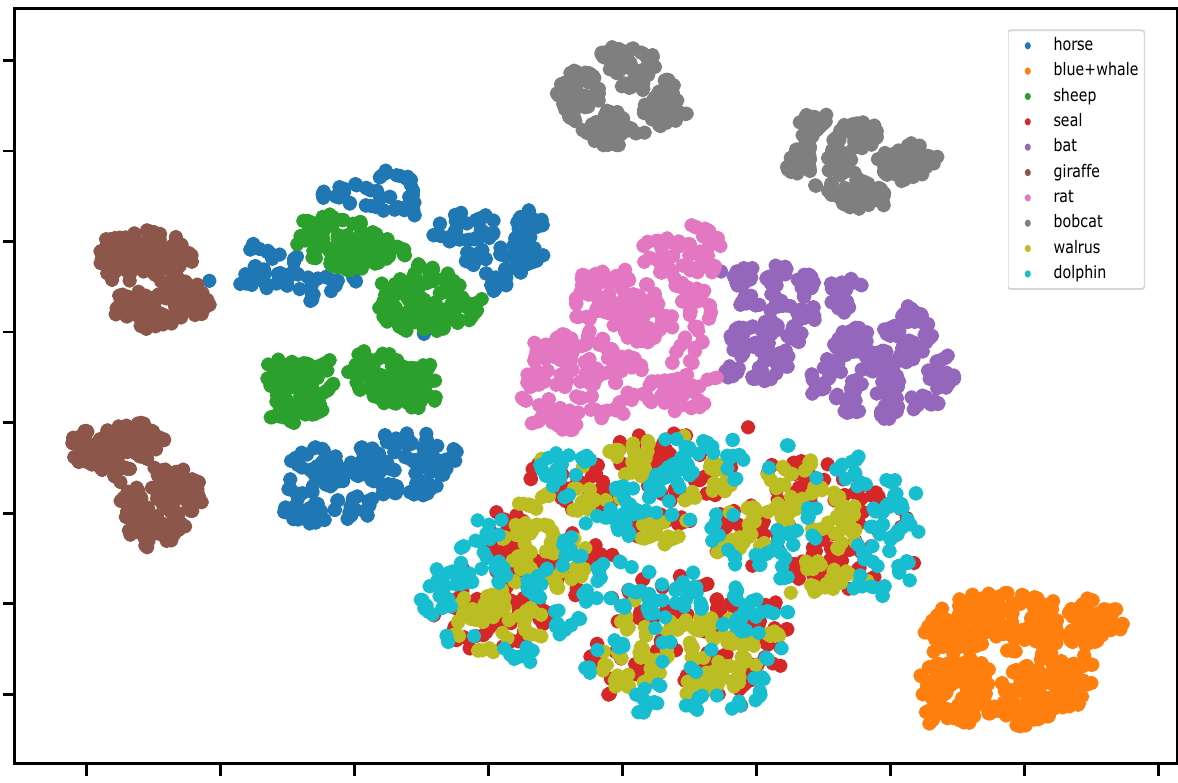}
\caption{AWA2 by ZSML. }
\end{subfigure}\hfil 
\begin{subfigure}{0.23\textwidth}
\centering
\includegraphics[width=\textwidth]{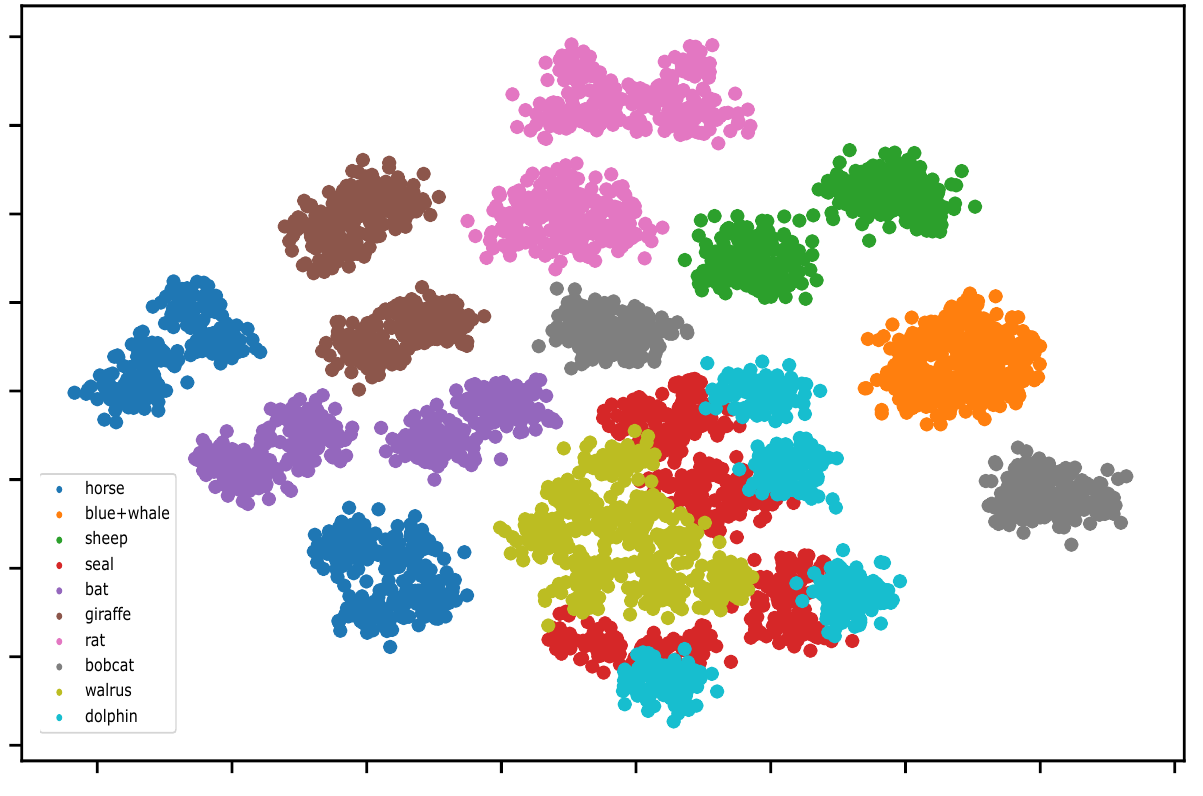}
\caption{AWA2 by AMAZ. }
\end{subfigure}\hfil 
\begin{subfigure}{0.23\textwidth}
\centering
\includegraphics[width=\textwidth]{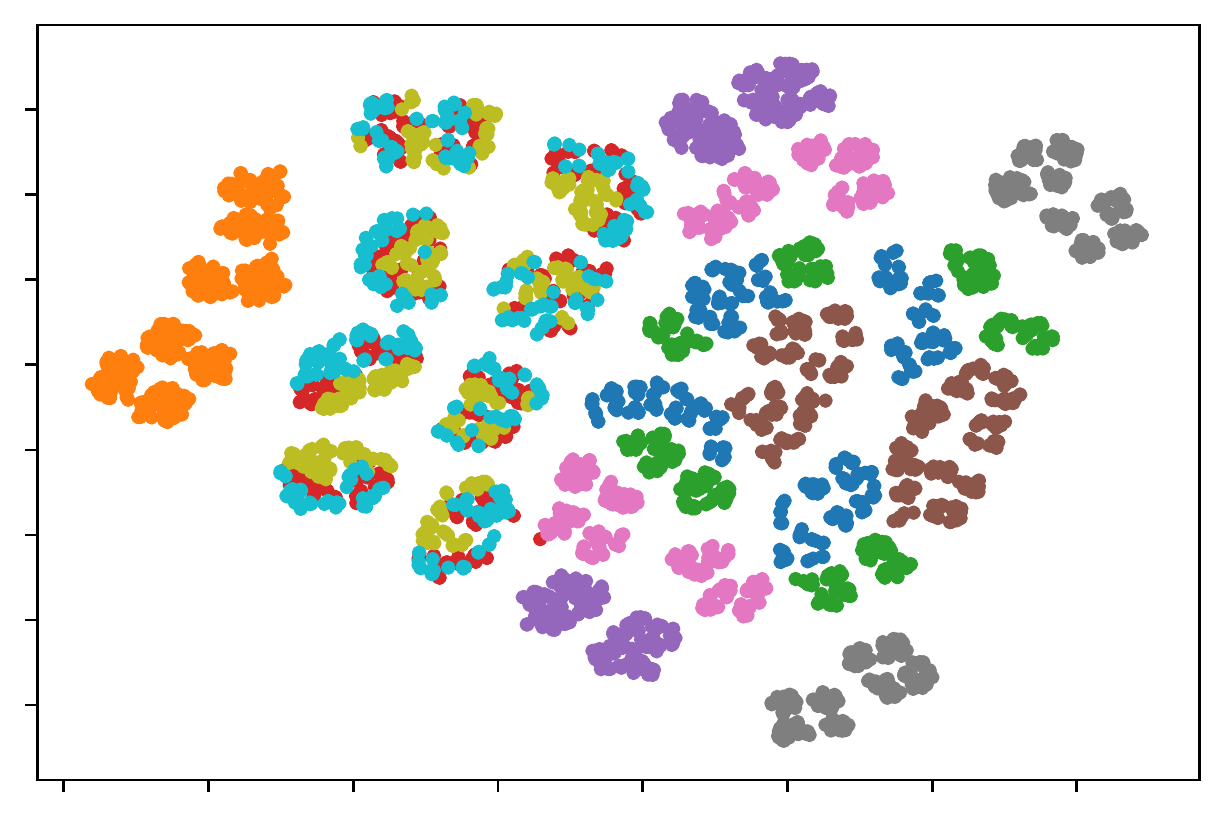}
\caption{AWA1 by ZSML. }
\end{subfigure}\hfil 
\begin{subfigure}{0.23\textwidth}
\centering
\includegraphics[width=\textwidth]{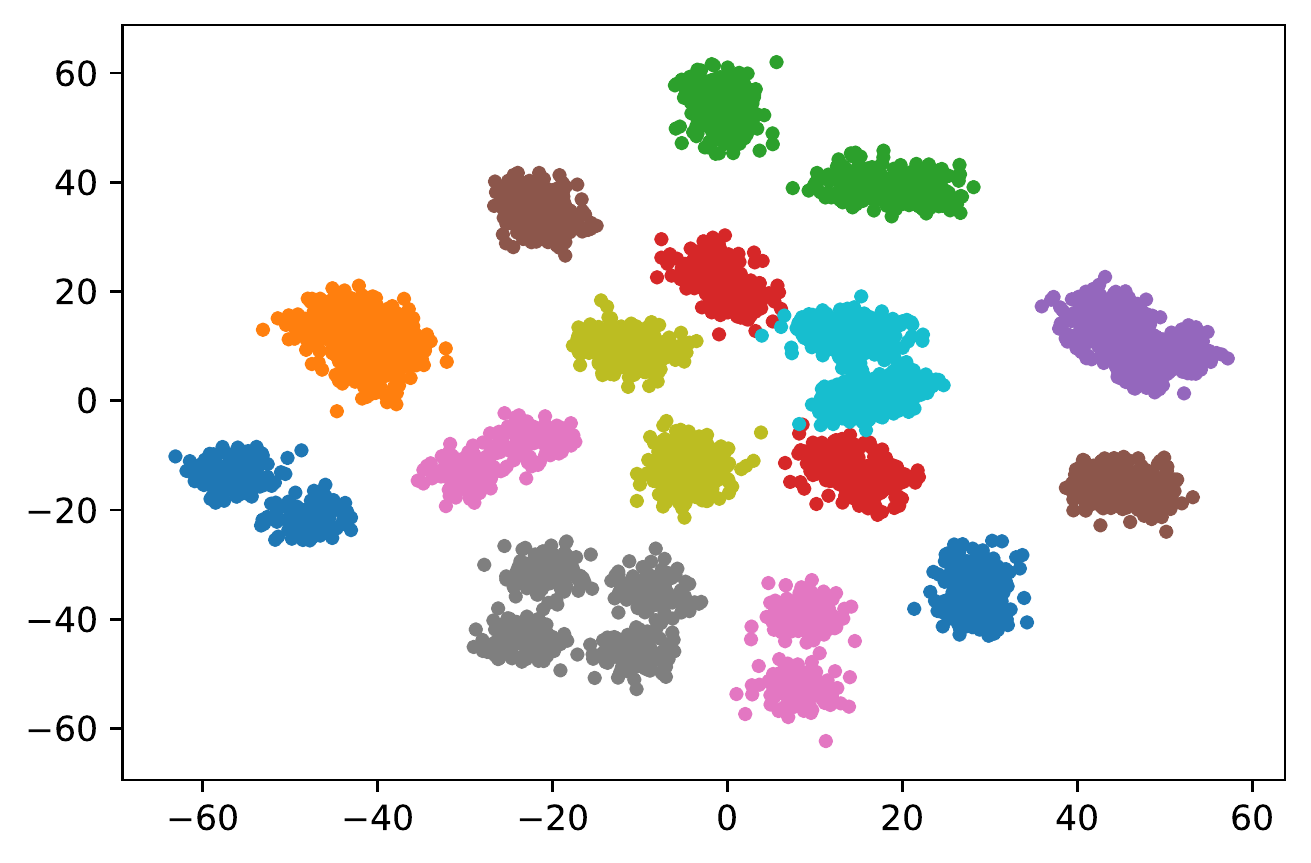}
\caption{AWA1 by AMAZ. }
\end{subfigure}\hfil 
\\
\begin{subfigure}{0.23\textwidth}
\centering
\includegraphics[width=\textwidth]{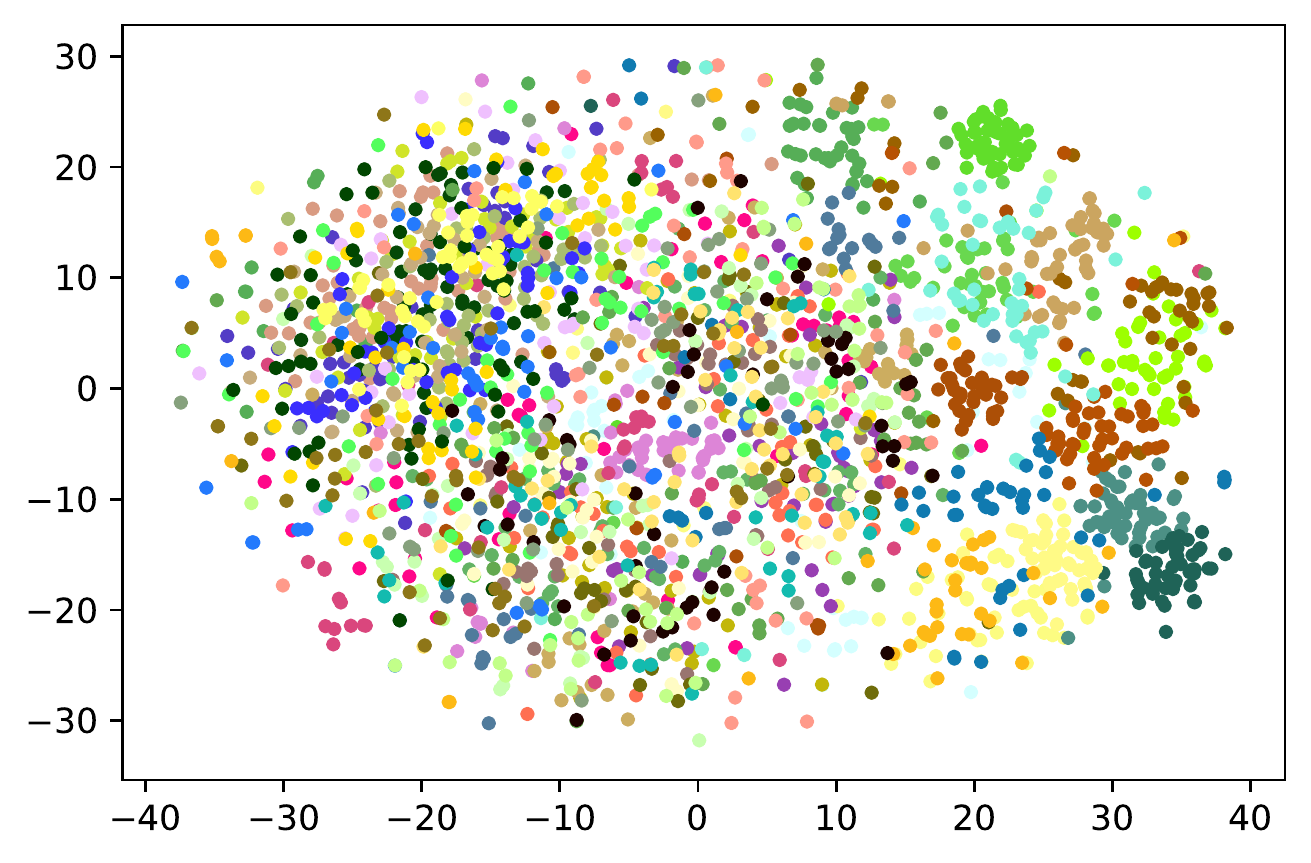}
\caption{CUB by ZSML. }
\end{subfigure}\hfil 
\begin{subfigure}{0.23\textwidth}
\centering
\includegraphics[width=\textwidth]{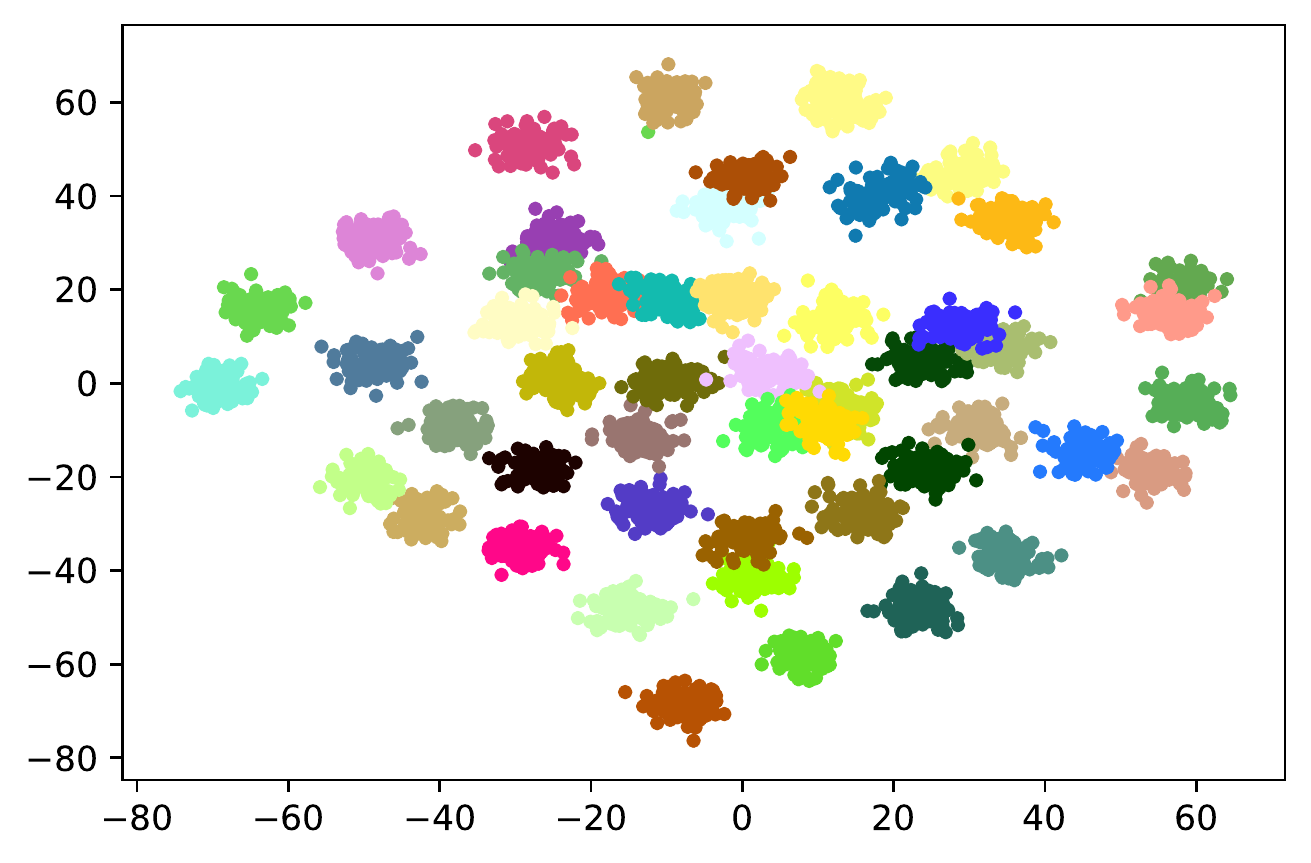}
\caption{CUB by AMAZ. }
\end{subfigure}\hfil 
\begin{subfigure}{0.23\textwidth}
\centering
\includegraphics[width=\textwidth]{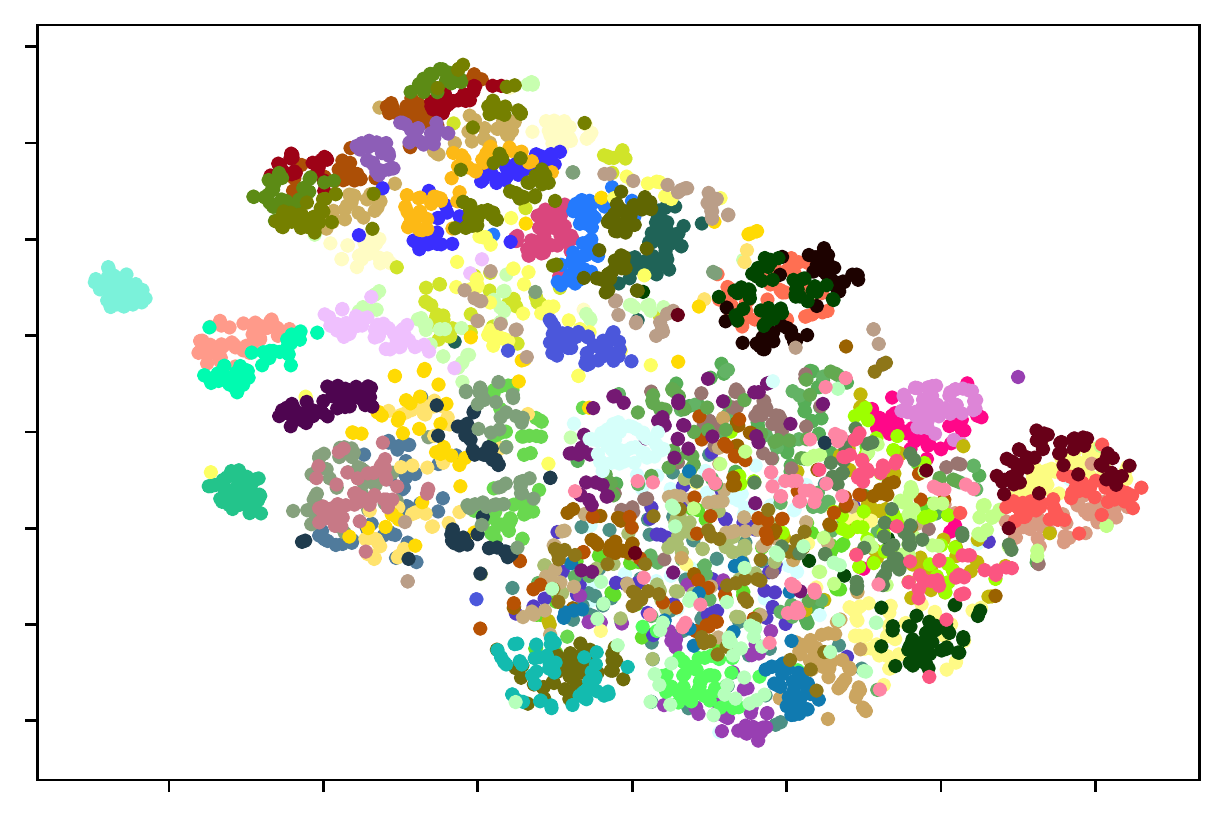}
\caption{SUN by ZSML. }
\end{subfigure}\hfil 
\begin{subfigure}{0.23\textwidth}
\centering
\includegraphics[width=\textwidth]{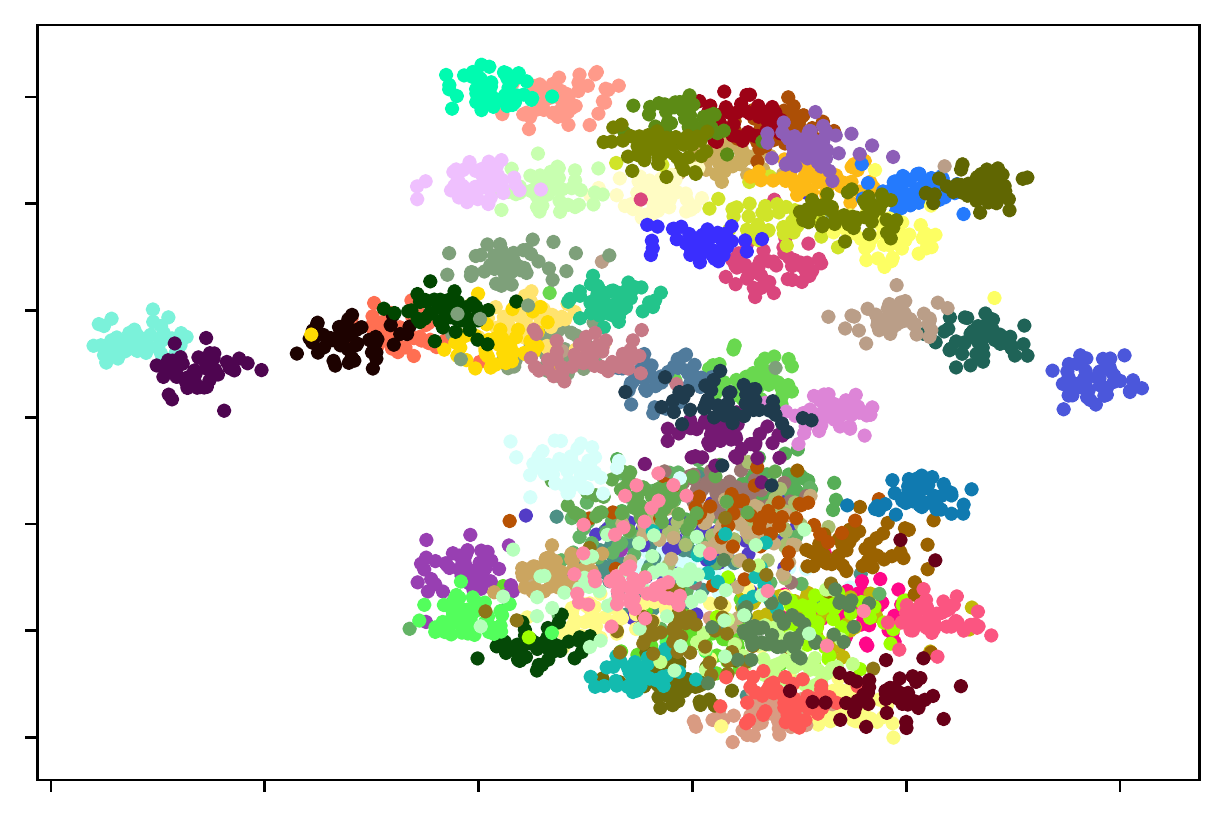}
\caption{SUN by AMAZ. }
\end{subfigure}\hfil 
\caption{Visualization of synthetic features on AWA2, AWA1, CUB, and SUN.}
\label{fig embedding}
\end{figure*}

\subsection{Modulator Operation Ablation Study}

In this section, we exhibit ablation study on how the attribute-aware modulator modulate the generator. Given the intermediate results $\{o_{j}\}_{j=0,1,...,k}$ of the generator and the attribute-aware parameters $\{(w_{j},b_{j})\}_{j=0,1,...,k}$, the intuitive baseline design of the modulator is $w_{j}o_{j}$. Based on the baseline, we further consider $o_j$ as the \textit{base}, $b_j$ as the \textit{bias}, activation function on $w_j$ or $b_j$ as \textit{activation}, operator, e.g., $+$, that connecting different components as \textit{operator}. The ZSL results of adopting different modulator on AWA2 are shown in Table~\ref{table modular}. To avoid the influence of final classifier, we utilize SVM since its performance is more stable than Softmax. As shown in Table~\ref{table modular}, adding \textit{base} and \textit{bias}, using + as \textit{operator}, and Sigmoid as \textit{activation} are effective methods to improve the performance of the modulator.

\subsection{Feature Visualization}

We synthesize 500, 500, 100, and 50 features using ZSML and AMAZ for each unseen class in AWA2, AWA1, CUB, and SUN, respectively, and visualize the features with t-SNE. 
We chose the synthetic number to make sure the number of all synthetic features is around 5000. 
The generated features are visualized with t-SNE, and results are shown in Figure~\ref{fig embedding}. 
Legend of AWA1 is the same as AWA2. And for CUB and SUN, due to lack of space, legends are not plotted. 
In Figure~\ref{fig embedding}(a), the features of `seal', `dolphin', and `walrus' generated by ZSML highly overlap with each other, which makes sense since they are biologically similar.
However, in Figure~\ref{fig embedding}(b), these three unseen animals can be easily separated in our feature space, which validates our model's superiority. 
Also, for AWA1 and CUB datasets, as shown in Figure~\ref{fig embedding} (c) and (e), features generated by ZSML are highly overlapped for some classes, while features generated by AMAZ can be easily separated according to Figure~\ref{fig embedding} (d) and (f). Besides, for SUN, from Figure~\ref{fig embedding} (g) and (h), we can find that features generated by AMAZ are apparently easier to distinguish than features generated by ZSML.

Besides, as shown in Figure~\ref{fig embedding} (e) and (f), the difference between the two visualization results is remarkably evident on CUB. 
The visualization of features generated by ZSML are highly scattered. The visualization difference is significant, while the performance gap is relatively trivial (1.7\%). The underlying reason is that although we generated more discriminative samples, the generated samples may not reflect the real images of unseen classes due to domain shift. An example is that the attribute being ‘has tail’, the seen class being monkey, and the unseen class being horse; although our generated sample can be more discriminative via better reflecting ‘has tail’, the generated features for horse may still be insufficient for training since monkey and horse have tails of different appearance. 

\subsection{Hyper-parameter Analysis}

\begin{figure}[htb]
\centering 
\begin{subfigure}{0.23\textwidth}
\centering
\includegraphics[width=\textwidth]{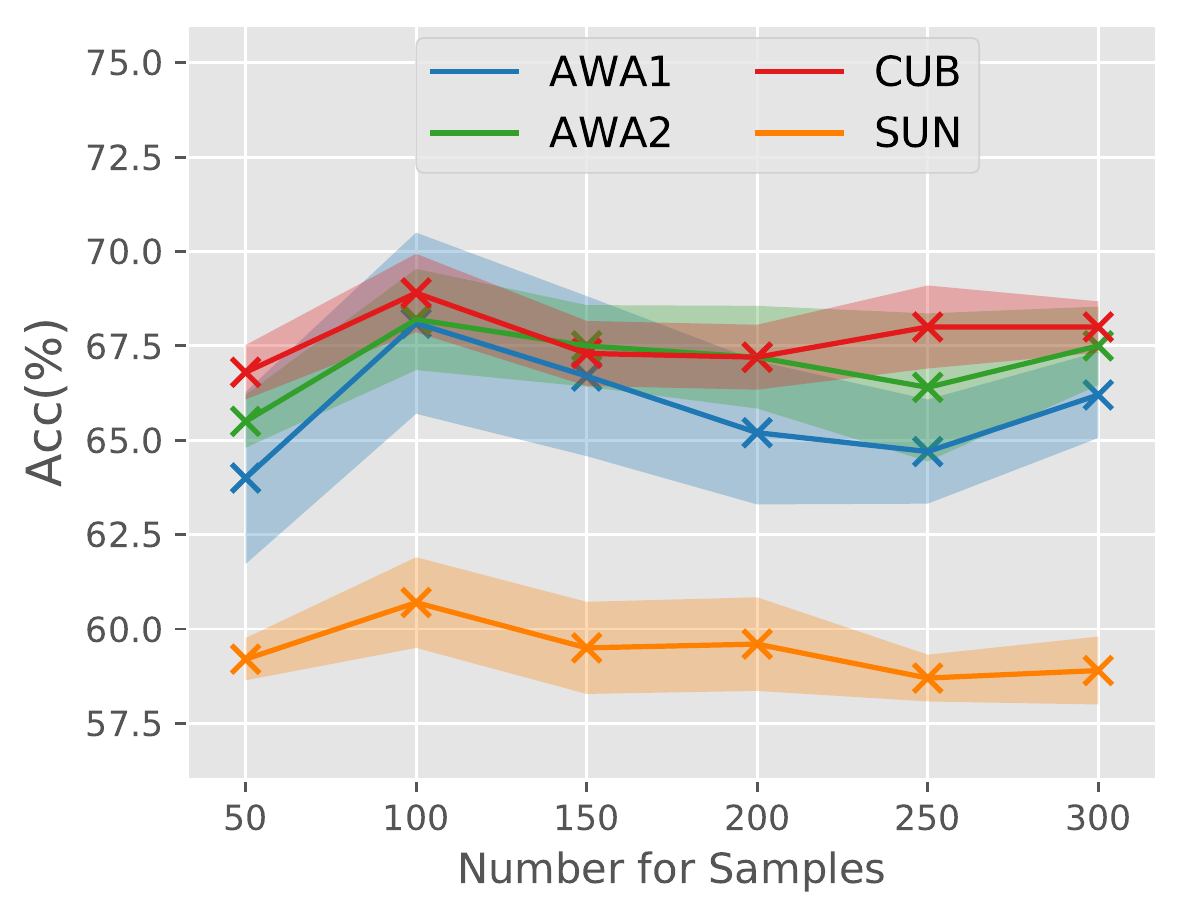}
\caption{Impact of feature \#. }
\end{subfigure}\hfil 
\begin{subfigure}{0.23\textwidth}
 \centering
\includegraphics[width=\textwidth]{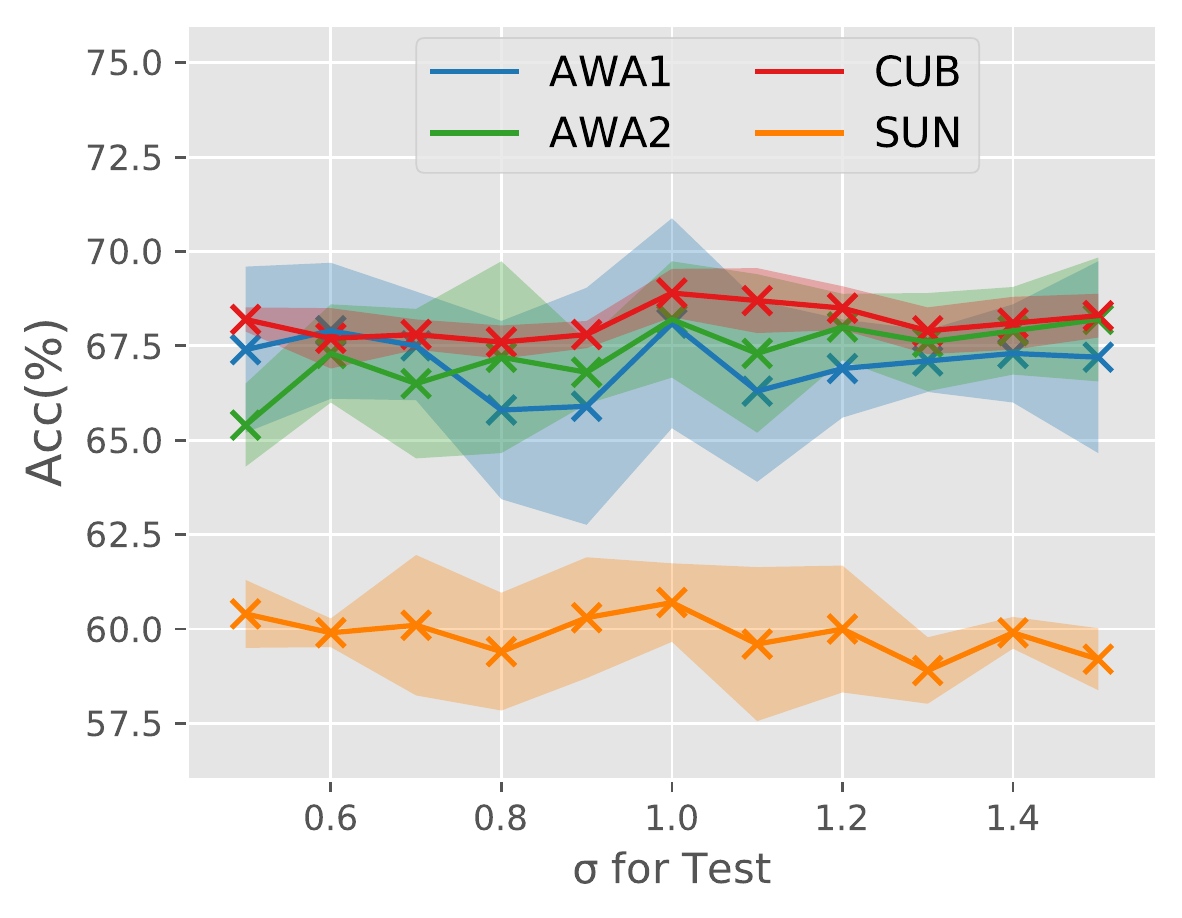}
\caption{Impact of $\sigma$. }
\end{subfigure}\hfil 
\caption{Hyper-parameter analysis in ZSL. Shadow alongside the curves represents standard deviation.}
\label{fig ablation}
\end{figure}

\begin{figure}[htb]
\centering 
\begin{subfigure}{0.235\textwidth}
\centering
\includegraphics[width=\textwidth]{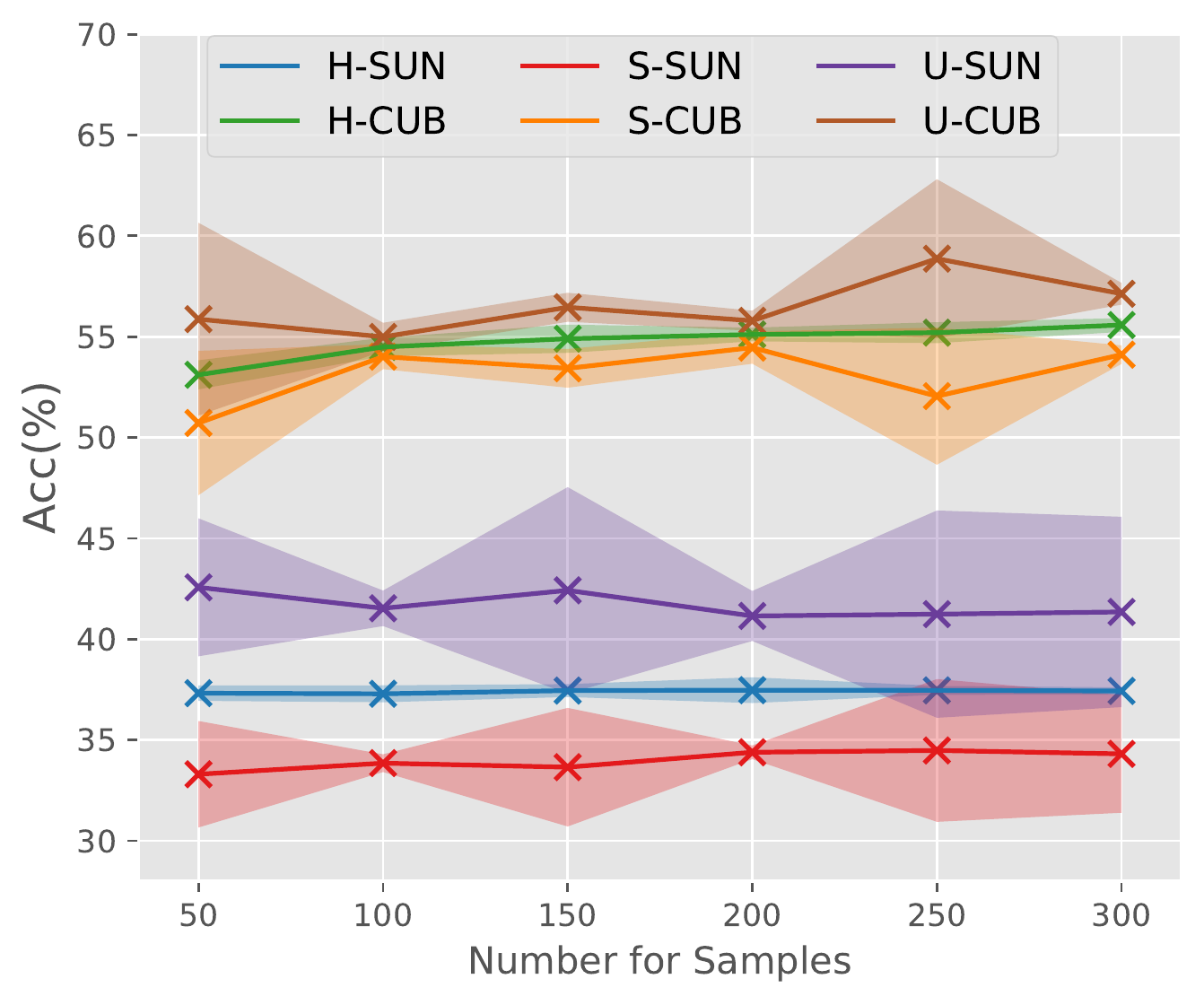}
\caption{Impact of feature \#. }
\end{subfigure}\hfil 
\begin{subfigure}{0.235\textwidth}
 \centering
\includegraphics[width=\textwidth]{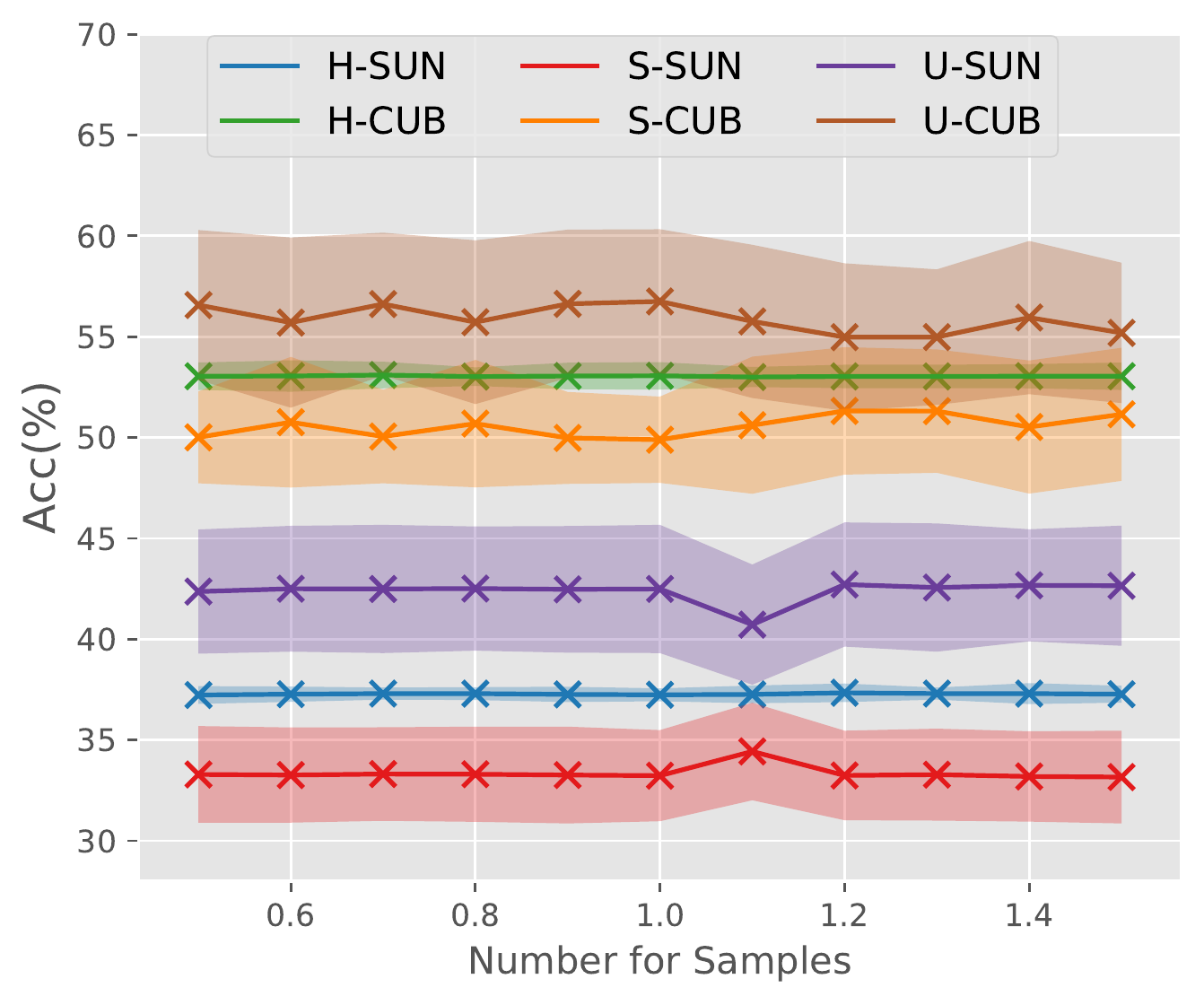}
\caption{Impact of $\sigma$. }
\end{subfigure}\hfil 
\caption{Hyper-parameter analysis in GZSL on CUB and SUN. Shadow alongside the curves represents standard deviation.}
\label{fig gzsl ablation}
\end{figure}

We conduct ablation studies to investigate how the number of synthetic samples and $\sigma$ of noise $z$ influence ZSL and GZSL. Experiments are done by our weighted-soft version classifier. 

\textbf{Zero-shot Learning}: The results in Figure~\ref{fig ablation}(a) and (b) show that AMAZ achieves robust performance, which changes slightly on the four datasets when parameter values increase. AMAZ achieves satisfactory results when $\sigma=1$ and sample number reaches 100.
The performance on AWA1 and AWA2 is more impacted by parameters, reflected by larger standard derivations in per-class accuracy. The reason is that seen and unseen classes are more different in AWA1 and AWA2, thus introducing a greater bias.

\textbf{Generalized Zero-Shot Learning}: We utilize real data for seen classes and synthetic features for unseen classes to train the classifier. The curves of the harmonic mean (H-SUN and H-CUB) in Figure~\ref{fig gzsl ablation} show that comparing with ZSL, the model performance is more stable when varying synthetic numbers and $sigma$. The reason is that classifier needs to predict labels for more classes on CUB and SUN in GZSL (50 vs. 200, and 72 vs. 717); thus only significant improvement or drop of the model performance can result in fluctuation of H. 


\section{Related Work}

\subsection{Zero-shot Learning}
A common strategy views zero-shot learning as an embedding problem of visual or semantic features.  
For example, Ye et al.~\cite{ye2019progressive} design a progressive ensemble network for learning a mapping function from the same extracted features to different label representations. \textcolor{black}{Bendre et al.~\cite{bendre2021generalized} propose multi-model cariational autoencoder based on a multi-modal loss to correlate modalities and global-local semantic knowledge for ZSL.}
Han et al.~\cite{han2020learning} utilize the mutual information to learn the redundancy-free visual embedding for better discrimination of features.
Imrattanatra et al.~\cite{imrattanatrai2019identifying} propose an embedding model based on a knowledge graph.
Such methods resort to learning a projection from visual space to semantic space~\cite{ye2019progressive,sung2018learning,kodirov2017semantic,LuLNCZ21,imrattanatrai2019identifying,li2020learning} or the reverse~\cite{zhang2017learning,shigeto2015ridge}. Then, ZSL can be accomplished by ranking similarity or compatibility in the shared space. 
Embedding can be learned based on the features extracted utilizing pre-trained backbones (e.g., ResNet101) or in an end-to-end manner~\cite{GuanSGLCN21}.
Zhang et al.~\cite{zhang2017learning} propose a multi-modality fusion method that enables end-to-end learning of semantic descriptors. \textcolor{black}{Moreover, Guo et al.~\cite{guo2017zero} propose a one-step recognition framework to perform recognition in the original feature space and thus avoid information loss of the intermediate transformation.} 

In the more challenging GZSL setting, where only instances from seen classes are provided, the embedding methods are more prone to suffer from data imbalance in recognizing data from both seen and unseen classes.
Conventional embedding methods
cannot address such problems well~\cite{xian2019zero}.

To address the data imbalance issue, several others efforts \cite{wang2018zero,zhang2019triple,gao2020zero,yang2020simple,xu2020correlated,yang2018imagination,gune2020generalized,YanCLGGZZ21} explore generative methods for ZSL. Felix et al.~\cite{felix2018multi} use a generative adversarial network to synthesize features constrained by multi-modal cycle-consistent semantic compatibility. 
Variational autoencoder\cite{mishra2018generative,schonfeld2019generalized,ma2020variational} is also adopted to avoid mode collapse caused by the structure of GAN. 
The f-CLSWGAN~\cite{xian2018feature} synthesizes the unseen instances according to the semantic descriptions.
Zero-VAE-GAN~\cite{gao2020zero} combines two generative models, variational autoencoder (VAE) and generative adversarial networks (GAN), to improve the model's performance and robustness.
Xu et al.~\cite{xu2020correlated} adopt two couple Wassertein GANs to generate semantic-related multi-modal features for further image retrieval.
By generating samples for unseen classes, ZSL is converted to supervised classification, and conventional classification methods can be applied.
In our model, we propose an attribute-weighted loss to enhance both deep learning and traditional machine learning classifiers.

Recently, meta-model is proposed to further eliminate the bias towards seen classes in ZSL~\cite{verma2020meta,meta_zeroshot1,meta_zeroshot2,meta_zeroshot3,meta_zeroshot4,liu2021task}. For example,
\textcolor{black}{Yu et al.~\cite{yu2020episode} first introduce an episode-based training framework for ZSL, which can progressively accumulate ensemble experiences based on the mimetic unseen classes and thus generalize the semantic prototypes for real unseen classes.} TAFE-NET~\cite{wang2019tafe} uses a meta learner for task-aware feature embedding. ZSML~\cite{verma2020meta} \textcolor{black}{introduce and meta-learning and generative network in ZSL.} \textcolor{black}{Liu et al.~\cite{liu2021task} propose to utilize a task-wise attribute alignment network to mitigate the potential biased meta-learning. They all rely on MAML~\cite{finn2017model} and generative model}. 
Although meta-models have achieved great success, they seek a common solution to be shared across tasks and thus fail to accommodate diverse or new tasks.
To address this limitation, our model introduces a modulation network and promises attribute-aware meta-learning.

Considering that generative methods cannot guarantee the quality of generated data for unseen classes due to the absence of real images, transductive methods are proposed to address the limitation~\cite{wan2019transductive,gao2020zero,wan2019transductive,rahman2019transductive}. Differing from ZSL, transductive methods assume that unlabeled images from unseen classes may occur during training and use the unlabeled images as auxiliary information.
Since this assumption is not strict as ZSL, we adopt the traditional inductive ZSL yet improve its generating quality by employing the attribute-aware modulation network. The attribute modulator modifies the generator towards attribute-awareness and attribute-richness.

\subsection{Feature Modulation}
Our AMAZ is also related to feature modulation, which explores the modulation of fully connected networks or convolutional networks.
Some research~\cite{zhang2020deep,de2017modulating,perez2017film} introduces conditional batch normalization to modulate a target neural network's visual processing based on linguistic input. \textcolor{black}{For example, Yu et al.~\cite{yu2021knowledge} directly generate the classifiers based on the class descriptions and semantic information of target unseen classes. Li et al.~\cite{li2019zero} use the area under score curve as weights to adapt tasks of ZSL and thus learn characterized semantic concepts.}
Our attribute-aware modulation can be viewed as an episode-wise feature modulation conditioned on attributes.
But inspired by attention mechanism~\cite{zhang2019self}, our model applies gate value, e.g., Sigmoid, to generated parameters before modulation. The gate value helps emphasize more attribute-related features and thus removes redundancy from synthetic instances. Compared with previous work~\cite{zhang2020deep,de2017modulating,perez2017film}, our model is more suitable for zero-shot learning, as it modulates the generator to reflect the semantic characteristics.

\subsection{Model-Agnostic Meta-Learning}
\textcolor{black}{Model-Agnostic Meta-Learning (MAML)~\cite{finn2017model} is an meta-learning based optimization framework. It aims to find a initial model, which can be fast adapted to other tasks with few samples. The authors achieve the goal by designing a two-step optimization procedure: meta-training and mete-validation. During meta-training, MAML optimizes the model towards different directions to fit different tasks. Then, during meta-validation, MAML aggregates the gradients from all optimization directions in meta-training and minimizes the overall validation loss based on validation tasks. Our model follows the same training procedure of MAML except that we update parameters per batch instead of per task to improve robustness.}

\section{Conclusion}
In this paper, we present an attribute-modulated generative meta-model (AMAZ) to synthesize visual features for unseen classes for ZSL. 
AMAZ incorporates an attribute-aware modulation network to modify the generator according to task characteristics learned from attributes. 
It introduces an attribute discriminator to guide the direction of modulation. Then the customized generator can be tuned for each task and adapt to diverse tasks, including new tasks.
In addition, AMAZ is trained in an episode-wise meta manner to mitigate the inherent bias caused by the absence of unseen data during training.
We further improve the AMAZ by an attribute-weighted classifier, which can denoise low-quality synthetic data.
Extensive experiments on four widely-used benchmarks show that our model exceeds state-of-the-art methods in both ZSL and GZSL settings.
The qualitative and quantitative experiments in zero-shot image retrieval also show that AMAZ generates more discriminative features.
In the future, we will extend the model to address the insufficiency of labeled data for more complex retrieval tasks, such as fine-grained image retrieval and cross-media retrieval.

\bibliographystyle{IEEEtran}

\bibliography{bio}

\end{document}